\documentclass{article}
\usepackage[english]{babel}
\usepackage{amssymb}

\usepackage[letterpaper,top=2cm,bottom=2cm,left=3cm,right=3cm,marginparwidth=1.75cm]{geometry}

\usepackage{amsmath}
\usepackage{graphicx}
\usepackage{algorithm}
\usepackage{algpseudocode}
\usepackage{subcaption}
\usepackage[colorlinks=true, allcolors=blue]{hyperref}

\title{A Survey of Nonlinear Filtering Methods for Data Assimilation and Recent Advances in Generative AI-Enabled Score Filters}
\author{
  Cath Pham$^1$, Kadesha Reynolds$^1$\\[1em]
  {\small{$^1$ Department of Mathematics, Florida State University, Tallahassee, FL}} 
}

\date{}

\begin{document}
\maketitle

\begin{abstract}
Many real-world applications, such as numerical weather prediction or wildfire tracking, involve modeling highly nonlinear dynamical systems that are chaotic and influenced by various sources of uncertainty. To improve model prediction accuracy and reduce such uncertainties, data assimilation (DA) combines the dynamical model with partial, noisy observations to estimate the evolution of the system's state. Doing so depends on solving the optimal filtering problem: computing the conditional distribution of a hidden state given a sequence of observations. Classical approaches fail along three distinct axes: Kalman-based methods assume Gaussian posteriors and produce biased estimates under nonlinear observation operators; particle-based methods represent distributions with weighted samples and collapse in high dimensions due to the curse of dimensionality; and kernel-based methods move beyond discrete particle representations but cannot maintain their density approximation affordably as dimension grows. The Ensemble Score Filter (EnSF) avoids all three constraints by encoding the filtering density in a score function and approximating that score with a training-free Monte Carlo estimator, enabling stable filtering in high-dimensional, chaotic systems. This survey traces EnSF's mathematical foundations, establishes it as the state of the art for non-Gaussian, high-dimensional filtering, and examines fourteen extensions that expand its operational reach. Twelve of the fourteen extensions preserve this training-free Monte Carlo estimator at the core of the update, with only two departing from it to achieve planetary-scale deployment.
\end{abstract}

\section{Introduction}
Accurate forecasting of chaotic physical systems, from predicting hurricane tracks and wildfire spread to calibrating global climate simulations, requires continuously tracking a hidden state from noisy, partial observations. This is the central task of data assimilation, which fuses observational data with dynamical models to maintain accurate estimates of evolving systems as new data arrives. Computing the best estimate exactly is analytically intractable in all but the simplest settings, so every practical method must decide how to represent the uncertainty in the state and how to update that representation as new observations arrive. These two choices determine what a method can do and where it structurally fails.

Two families of approximation methods have dominated the field for decades, and each achieves some of three properties a filtering method needs: non-Gaussian expressiveness, scalability to high dimensions, and affordability as posterior complexity grows. Kalman-based filters represent the filtering density as a Gaussian, which enables analytic updates that scale efficiently to very high dimensions and remain cheap to maintain regardless of posterior complexity \cite{evensen1994sequential, hunt2007efficient, kalman1961new, kalman1960new}, but this Gaussian representation fails structurally whenever the true posterior is non-Gaussian. Particle-based filters represent the density as a set of weighted samples, which captures arbitrary non-Gaussian structure \cite{doucet2001introduction, gordon1993novel}, but under high-dimensional, strongly coupled observations, the number of samples required grows exponentially with dimension, making particle filters both unscalable and unaffordable in practice \cite{snyder2008obstacles}. This is the ceiling that classical filtering could not escape, and it is the organizing question of this survey.

EnSF addresses this issue by building upon two ideas. The first is rooted in the theory of backward stochastic differential equations (BSDEs), originally developed in stochastic control and later brought into filtering \cite{bao2011numerical, pardoux1990adapted, pardoux1994backward}. This lineage established that running a stochastic process backward in time extracts the distributional information the filtering update requires, with a martingale correction term keeping the backward evolution uniquely solvable, and it culminated in the kernel mixture representation this survey adopts as its BSDE filter \cite{archibald2022kernel}. The second is rooted in score based generative models \cite{song2020score}, which showed how to generate samples from a target distribution by running a stochastic process backward in pseudo time instead of physical time. A forward diffusion process maps the target distribution to a simple reference distribution in finite pseudo time, typically a standard Gaussian, and the reverse process starts from that fixed reference and recovers the target. This removes the expensive problem specific terminal conditions that made filtering based on BSDEs computationally prohibitive \cite{bao2024score}.

The Ensemble Score Filter (EnSF) builds on these two insights into a practical filtering algorithm \cite{bao2024ensemble}. Rather than storing the filtering density as a parametric Gaussian or particles that directly represent the density, EnSF encodes it in the score function, which is the gradient of the log-density with respect to the state. EnSF uses a reverse-time stochastic differential equation guided by this score to generate posterior samples at each assimilation step. Importantly, EnSF approximates the score through a closed-form training-free Monte Carlo estimator that requires no neural network retraining between steps, resolving the primary computational bottleneck of earlier score-based filters and enabling stable filtering in very high dimensional systems, such as those with one million state dimensions \cite{bao2024ensemble}.

This survey traces the mathematical foundations and development of EnSF and examines the framework it has generated. Following a critical analysis of classical methods, the paper examines fourteen extensions of EnSF, organized as five acts that test whether EnSF's core mechanism, once established, can absorb escalating operational demands without needing to change. The acts progress from improving the algorithm itself to testing whether the improved mechanism holds on harder states and forecasting problems. They then reach the point where the mechanism requires modification, and finally where its training-free property is given up for planetary-scale deployment. By characterizing every method according to how it represents and updates the filtering density, the survey establishes EnSF as the state of the art for non-Gaussian high-dimensional filtering, escaping the tradeoff between non-Gaussian expressiveness, scalability, and affordable density representation that constrained every classical approach.

\section{Mathematical Preliminaries}
\subsection{Optimal Filtering}
The optimal filtering problem is rigorously formulated as a state-space model consisting of a hidden Markov state process and a noisy observation process \cite{jazwinski2007stochastic}.

Let $t \in \mathbb{Z}^+$ represent the discrete time index. The unobservable dynamical state $X_t \in \mathbb{R}^d$ is controlled by the nonlinear forward model
\begin{equation} \label{eq:state}
    \text{State: } X_{t+1} = f(X_t,\omega_t),
\end{equation}
where $f: \mathbb{R}^d \times \mathbb{R}^k \mapsto \mathbb{R}^d$ defines the physical dynamics, and $\omega_t \in \mathbb{R}^k$ represents the random model uncertainty or the incomplete knowledge of its underlying physics. Since there is no direct access to the real state, inference must be drawn from a series of partial, noisy measurements. This observation process is defined as 
\begin{equation} \label{eq:obs}
    \text{Observation: } Y_{t+1} = g(X_{t+1}) + \epsilon_{t+1},
\end{equation}
where $Y_{t+1} \in \mathbb{R}^r$ is the observational data, $g$ is the possibly nonlinear observation operator, and $\epsilon_{t+1} \sim \mathcal{N}(0, \Sigma)$ denotes observational Gaussian noise, with $\Sigma \in \mathbb{R}^{r \times r}$ is the observation noise covariance matrix. 

The ultimate goal of optimal filtering is to compute the best estimate of the hidden state given all available observations. Mathematically, this requires finding the optimal estimate $\hat{X}_{t+1}$, defined as the conditional expectation of the state $\hat{X}_{t+1} := \mathbb{E}[X_{t+1}|\mathcal{Y}_{t+1}]$, given the complete history of the observation $\mathcal{Y}_{t+1} := \sigma(Y_{1:t+1})$. The evolving sequence of $\sigma$-algebra forms a filtration that encodes the accumulation of observational information over time, ensuring that the state estimate at each step depends only on observations up to that moment \cite{jazwinski2007stochastic}. As new data accumulates over time, the estimation accuracy progressively refines.

Computing this conditional expectation directly is generally intractable for high-dimensional problems. A more general approach is instead to characterize the full conditional distribution $P(X_{t+1}|\mathcal{Y}_{t+1})$, known as the \textit{filtering density}, from which various statistical quantities, including the conditional mean and covariance, can be derived. The following section introduces the Bayesian filtering framework used to characterize the evolution of this conditional distribution \cite{kallianpur1968estimation, stratonovich1965conditional, zakai1969optimal}.

\subsection{Recursive Bayesian Filter}
The recursive Bayesian filter estimates the optimal filtering density over time using a two-step loop: prediction step and update step. This mathematical framework is the foundation of the traditional estimation method \cite{jazwinski2007stochastic}. To align with the mathematical notation in later equations, this paper adopts specific formulations in the score-based approach research \cite{bao2024score}.

Assume the posterior filtering density $P(X_t \mid \mathcal{Y}_t)$ is available at time $t$. The goal of the prediction step is to propagate the posterior filtering density forward from time $t$ to $t+1$ to obtain the prior filtering density. It uses the Chapman-Kolmogorov equation:
\begin{equation} \label{eq:prior_chapman}
    \text{Prior: } P(X_{t+1}|\mathcal{Y}_t) = \int P(X_{t+1}|X_t)P(X_{t}|\mathcal{Y}_t)dX_t,
\end{equation}
where $P(X_{t+1}|X_t)$ is the stochastic transition probability derived from the state (Eq.~\ref{eq:state}), $P(X_{t}|\mathcal{Y}_t)$ is the posterior filtering density from time $t$.

The goal of the update step is to assimilate the new observational data $Y_{t+1}$ (the likelihood) with the prior estimate to get the posterior filtering density. The Bayes' theorem is the mechanism that connects the prior filtering density to the posterior filtering density at time $t+1$.
\begin{equation} \label{eq:posterior}
    \text{Posterior: } P(X_{t+1}|\mathcal{Y}_{t+1}) \propto P(X_{t+1}|\mathcal{Y}_t)P(Y_{t+1}|X_{t+1}),
\end{equation}
where the likelihood $P(Y_{t+1}|X_{t+1})$ is defined assuming Gaussian observation noise $\epsilon$ with its covariance matrix $\Sigma$:
\begin{equation} \label{eq:likelihood}
    P(Y_{t+1}|X_{t+1}) \propto \exp[-\frac{1}{2}(g(X_{t+1})-Y_{t+1})^{\top}\Sigma^{-1} (g(X_{t+1})-Y_{t+1})]
\end{equation}

In practice, neither the Chapman-Kolmogorov integral (Eq.~\ref{eq:prior_chapman}) nor the Bayesian update (Eq.~\ref{eq:posterior}) can be computed exactly. The integrals are analytically intractable whenever $f$ or $g$ are nonlinear, and numerical evaluation becomes exponentially expensive as dimension grows. The following section examines three classical strategies that each resolve this intractability in a different way.

\section{Classical methods}
Practical filtering methods differ in how they represent the filtering distribution and how they update this representation when new observations become available. This section examines three representative approaches based on these two aspects. The Kalman filter, the particle filter, and the BSDE filter each arose from a distinct mathematical tradition, namely control theory, sequential Monte Carlo, and stochastic analysis respectively, and each independently revealed a different facet of the same underlying difficulty. Despite their successes, these methods face different limitations when applied to nonlinear, high-dimensional systems, particularly in representing complex non-Gaussian distributions and maintaining computational efficiency.

\subsection{Kalman filter}
The Kalman filter provides the exact solution to the optimal filtering problem under two assumptions: the state dynamics and observation operator must both be linear, and all noise must be Gaussian. Under linearity, a Gaussian state remains Gaussian after propagation through the model, so the filtering density stays within the Gaussian family at every time step. This closure property is what makes the Kalman filter analytically tractable. Rather than tracking an infinite-dimensional function $P(X_t|\mathcal{Y}_t)$, the filter maintains only the mean $\hat{X}_t$ and covariance $P_t$, which together fully characterize the Gaussian filtering density \cite{kalman1961new, kalman1960new}.

The linear state and observation equations are defined as
\begin{equation} \label{eq:kf_state}
    X_{t+1} = AX_t + CU_t, \quad U_t \sim \mathcal{N}(0, I)
\end{equation}
\begin{equation} \label{eq:kf_obs}
    Y_{t+1} = BX_{t+1} + DV_t, \quad V_t \sim \mathcal{N}(0, I)
\end{equation}
where $A \in \mathbb{R}^{d \times d}$ is the state transition matrix, $B \in \mathbb{R}^{r \times d}$ is the observation matrix, $C$ and $D$ are noise scaling matrices, and $U_t$, $V_t$ are independent standard Gaussian noise vectors. The filter tracks the conditional mean
\begin{equation} \label{eq:kf_mean}
    \hat{X}_t := \mathbb{E}[X_t | \mathcal{Y}_t]
\end{equation}
and the conditional error covariance
\begin{equation} \label{eq:kf_cov}
    P_t := \mathbb{E}[(X_t - \hat{X}_t)(X_t - \hat{X}_t)^\top]
\end{equation}

The filter proceeds through the recursive Bayesian loop in two steps. In the prediction step, the state mean and error covariance are propagated forward:
\begin{equation} \label{eq:kf_pred_mean}
    \hat{X}_{t+1|t} = A\hat{X}_t
\end{equation}
\begin{equation} \label{eq:kf_pred_cov}
    P_{t+1|t} = AP_tA^\top + CC^\top
\end{equation}
Equation~\ref{eq:kf_pred_cov} is the discrete Riccati equation. Because the dynamics are linear, this propagation is deterministic and exact. The filter then computes the Kalman gain
\begin{equation} \label{eq:kf_gain}
    K_{t+1} = P_{t+1|t}B^\top
    (BP_{t+1|t}B^\top + DD^\top)^{-1}
\end{equation}
which determines the relative weight given to the model prediction versus the new observation. When observation noise $DD^\top$ is large, $K_{t+1}$ is small and the filter weights the model prediction more heavily; when the predicted uncertainty $P_{t+1|t}$ is large, $K_{t+1}$ increases and the filter weights new observations more heavily.

In the update step, the mean and covariance are corrected using the incoming observation $Y_{t+1}$:
\begin{equation} \label{eq:kf_update_mean}
    \hat{X}_{t+1} = \hat{X}_{t+1|t} + 
    K_{t+1}(Y_{t+1} - B\hat{X}_{t+1|t})
\end{equation}
\begin{equation} \label{eq:kf_update_cov}
    P_{t+1} = (I - K_{t+1}B)P_{t+1|t}
\end{equation}
where $I$ is the identity matrix. The term $Y_{t+1} - B\hat{X}_{t+1|t}$ is the \textit{innovation}, which measures the discrepancy between the actual observation and the model's predicted observation.

The Kalman filter admits a closed-form analytic solution precisely because the Gaussian assumption is closed under linear transformations. When the dynamics or observation operator are nonlinear, propagating a Gaussian through a nonlinear function produces a non-Gaussian distribution that can no longer be exactly represented by a mean and covariance alone. The Riccati equation loses its deterministic structure, and the filter produces biased estimates \cite{jazwinski2007stochastic}. This limitation motivates a sequence of extensions, each of which attempts to extend the Kalman framework to progressively more nonlinear and high-dimensional settings while preserving its computational tractability.

The Extended Kalman Filter (EKF) linearizes the state and observation functions at each time step using a first-order Taylor expansion, producing a local Jacobian approximation around the current mean and then applying the standard Kalman update to this linearized model \cite{jazwinski2007stochastic}. When the true nonlinearity is mild, this approximation is adequate. Under strong curvature, the linear approximation degrades and the filter can diverge, since the Jacobian captures only local behavior and provides no information about the distribution's tails or multimodal structure \cite{jazwinski2007stochastic}.

The Unscented Kalman Filter (UKF) addresses this by propagating a deterministic set of $2d+1$ sigma points through the true nonlinear function rather than linearizing it \cite{julier1997new}. This captures the mean and covariance to second-order accuracy without requiring a Jacobian. Since the number of sigma points grows linearly with the state dimension, and the method also requires operations involving the full covariance matrix, the computational cost becomes prohibitive for very high-dimensional systems. More fundamentally, UKF still fits a Gaussian to the propagated sigma points, so any true multimodality or heavy-tailed structure in the posterior cannot be fully represented \cite{van2000unscented}.

The Ensemble Kalman Filter (EnKF) addresses the dimensionality problem by replacing the full $d \times d$ covariance matrix with a sample covariance estimated from an ensemble of $N \ll d$ model states, which has rank at most $N-1$ \cite{evensen1994sequential}. The Kalman update is applied to each ensemble member using this low-rank covariance, with the ensemble mean as the state estimate and the ensemble spread as the uncertainty. This is feasible because atmospheric error covariance is dominated by a limited number of large-scale structures, so a modest ensemble can capture its leading directions. However, the effective dimension of the full state still exceeds $N$, so the rank-deficient sample covariance is noisy and introduces spurious correlations between spatially distant and physically unrelated variables. These correlations degrade the posterior covariance estimate and can lead to filter divergence when they enter the update \cite{hamill2001distance, hunt2007efficient, whitaker2002ensemble}.

The Local Ensemble Transform Kalman Filter (LETKF) mitigates this by restricting each grid point's update to observations within a tunable localization radius, eliminating spurious long-range correlations by construction \cite{hunt2007efficient}. Rather than updating all variables simultaneously, LETKF solves a small $N \times N$ eigenvalue problem independently at each grid point, which has two important results. First, the localization eliminates the spurious correlations that afflict EnKF. Second, since the local analyses can be performed independently, LETKF is well suited to parallel computing architectures, enabling efficient implementation on large-scale computing systems. This combination makes LETKF the current operational standard at major weather forecasting centers, such as the Japan Meteorological Agency \cite{hunt2007efficient}. Its remaining limitations are that the localization radius is a manual hyperparameter requiring retuning for each new system, and more fundamentally, LETKF inherits the Gaussian assumption shared by the entire Kalman family \cite{hunt2007efficient}. Under strongly nonlinear observation operators, the linear ensemble update produces biased estimates regardless of how carefully the localization is tuned \cite{kurosawa2023statistical}.

Within this survey, the Kalman family represents the filtering density as a Gaussian characterized by a mean and covariance, and updates it through a linear correction proportional to the innovation. This representational choice enables exact analytic updates that scale efficiently to very high dimensions and maintain an affordable fixed-size parametric form regardless of posterior complexity. In the three-property framework of this survey, the Kalman family achieves both scalability and affordable density representation but sacrifices non-Gaussian expressiveness entirely. This ceiling motivates the development of methods that represent the filtering density without any parametric distributional assumption.

\subsection{Particle Filter}
Rather than representing the filtering density as a Gaussian, the particle filter represents it as a weighted empirical measure over a finite set of $M$ samples, called particles \cite{gordon1993novel}. This representation places no distributional assumption on the posterior and can therefore approximate any shape of filtering density, whether multimodal, skewed, or heavy-tailed. In the terminology of this survey, the particle filter stores the filtering density as a weighted empirical measure over discrete samples and updates it by reweighting those samples according to their likelihood under each new observation. This section adopts the formulations established in the score-based filtering framework \cite{bao2024score}.

The general weighted empirical approximation of the filtering density at time $t$ is
\begin{equation} \label{eq:pf_general}
    P(X_t|\mathcal{Y}_t) \approx 
    \sum^M_{m=1} w_{t,m} \, \delta_{x_{t,m}}(X_t)
\end{equation}
where $w_{t,m} \geq 0$ are normalized importance weights satisfying $\sum_m w_{t,m} = 1$, and $\delta_{x_{t,m}}$ denotes the Dirac delta mass at particle position $x_{t,m}$. After resampling, all weights are reset to $1/M$, recovering the equally weighted empirical measure
\begin{equation} \label{eq:pf_equal}
    P(X_t|\mathcal{Y}_t) \approx 
    \frac{1}{M}\sum^M_{m=1}\delta_{x_{t,m}}(X_t)
\end{equation}

The filter proceeds through the recursive Bayesian loop. In the prediction step, each particle is propagated independently through the state equation to obtain a set of predicted particles:
\begin{equation} \label{eq:pf_predict}
    \tilde{x}_{t+1,m} = f(x_{t,m}, \omega_{t,m})
\end{equation}
yielding the prior empirical approximation
\begin{equation} \label{eq:pf_prior}
    P(X_{t+1}|\mathcal{Y}_t) \approx 
    \frac{1}{M}\sum^M_{m=1}
    \delta_{\tilde{x}_{t+1,m}}(X_{t+1}).
\end{equation}
This is the Monte Carlo approximation of the Chapman-Kolmogorov integral (Eq.~\ref{eq:prior_chapman}).

In the update step, when the new observation $Y_{t+1}$ arrives, each particle is assigned a likelihood weight proportional to how well it explains the observation:
\begin{equation} \label{eq:pf_weights}
    w_{t+1,m} \propto P(Y_{t+1}|\tilde{x}_{t+1,m})
\end{equation}
with weights normalized to sum to one. The weighted posterior approximation is then 

\begin{equation} \label{eq:pf_posterior}
    P(X_{t+1}|\mathcal{Y}_{t+1}) \approx 
    \sum^M_{m=1} w_{t+1,m} \, 
    \delta_{\tilde{x}_{t+1,m}}(X_{t+1})
\end{equation}

To address weight degeneracy, in which likelihood weights concentrate on a small number of particles and leave the remainder with negligible weight, the particle filter incorporates a resampling step. The effective sample size (ESS)
\begin{equation} \label{eq:ess}
    \mathrm{ESS} = \frac{1}{\sum^M_{m=1} w_{t+1,m}^2}
\end{equation}
quantifies how many particles contribute meaningfully to the approximation. A common heuristic triggers resampling when the ESS falls below $M/2$, though the threshold is problem-dependent \cite{doucet2001introduction}. When resampling is triggered, $M$ new particles are drawn with replacement from the current weighted set, duplicating particles with large weights and discarding those with negligible weights, and all weights are reset to $1/M$:
\begin{equation} \label{eq:pf_resample}
    P(X_{t+1}|\mathcal{Y}_{t+1}) \approx 
    \frac{1}{M}\sum^M_{m=1}
    \delta_{x_{t+1,m}}(X_{t+1}).
\end{equation}

The particle filter provides the most general solution to the nonlinear filtering problem in low-to-moderate dimensions, where its non-Gaussian representation is most valuable. Under high-dimensional, strongly coupled observation scenarios, the filtering density becomes increasingly concentrated in a small region of the state space that a fixed particle budget cannot adequately cover \cite{snyder2008obstacles}. After the update step, almost all particles receive negligible likelihood weight under the observation, causing the ESS to collapse toward one. Resampling cannot recover diversity once weight collapse has occurred, because all resampled particles originate from the single surviving high-weight particle \cite{snyder2008obstacles}. In the three-property framework of this survey, the particle filter achieves non-Gaussian expressiveness but sacrifices both scalability and affordable density representation, since the two pressures compound: a complex posterior in high dimensions requires exponentially more particles to cover both the larger space and the more intricate shape simultaneously \cite{snyder2008obstacles}, making the particle filter computationally infeasible for the high-dimensional systems encountered in operational DA.

In the past two decades, several enhancements have been proposed to extend the regime in which the particle filter remains useful, each targeting a distinct failure mode. The Auxiliary Particle Filter (APF) addresses one-step weight collapse that occurs when the likelihood is sharply peaked relative to the prior \cite{pitt1999filtering}. Before propagation, APF estimates how compatible each particle's likely future position is with the upcoming observation $Y_{t+1}$, then reweights and resamples based on this look-ahead compatibility. This prevents particles from being propagated into regions where the likelihood will immediately be near zero, reducing degeneracy under highly informative observations. However, the look-ahead step requires a likelihood approximation that is not always accurate where the dynamics are diffuse relative to the likelihood \cite{andrieu2003efficient}.

The Rao-Blackwellized Particle Filter (RBPF) targets a different inefficiency: when the state can be partitioned into a nonlinear component and a conditionally linear component, applying a particle filter to the full state wastes particles on dimensions that admit an exact Kalman filter solution \cite{murphy2001rao}. RBPF tracks the nonlinear component with particles and the conditionally linear component with a Kalman filter conditioned on each particle, so the effective particle dimensionality is reduced to the nonlinear subspace. For systems where the nonlinear component is low-dimensional but the linear component is high-dimensional, as in target tracking and robotics, this makes particle use more efficient. When the state cannot be partitioned this way, RBPF offers no advantage over the standard particle filter \cite{murphy2001rao}.

The Regularized Particle Filter (RPF) addresses sample impoverishment, in which repeated resampling collapses the particle set to a small number of distinct positions \cite{arulampalam2002tutorial, musso2001improving}. After resampling, RPF perturbs each particle with kernel noise scaled by the empirical covariance of the ensemble, converting duplicate clusters into clouds of distinct nearby particles and restoring diversity. However, the bandwidth is only optimal for equally weighted Gaussian particles; it is a heuristic otherwise. Also, RPF changes only the resampling step, so the weight collapse in high dimensions remains \cite{arulampalam2002tutorial, snyder2008obstacles}.

None of these enhancements removes the underlying representational constraint. Storing the filtering density as a discrete weighted sample set requires exponentially many samples to maintain a faithful approximation as dimension grows. They extend the regime in which the particle filter is useful, but they do not change what the representation fundamentally is. This motivates a filtering density representation that does not rely on discrete weighted samples at all.

\subsection{Backward Stochastic Differential Equation (BSDE) Filter}
Kalman filters and particle filters approach the filtering density from opposite directions, yet each carries its own limitation: Kalman filters rely on linear and Gaussian assumption, while particle filters suffer from the curse of dimensionality. A third approach, rooted in the theory of backward stochastic differential equations (BSDEs), originally developed in stochastic control, represents the filtering density differently from both.

A numerical scheme for the forward-backward doubly stochastic differential equation (FBDSDE) system was developed to approximate the solution of the associated stochastic partial differential equation and its spatial gradient. The gradient provides additional information for computing quantities, such as the diffusion flux \cite{bao2011numerical}. This approach then extended to nonlinear filtering, where the BDSDE solution represents the unnormalized filtering density, which is normalized through the Kallianpur-Striebel formula. This filter outperformed both the particle filter and the Zakai filter, though the doubly stochastic term still required a specialized two-sided It\^o formula that made it cumbersome to solve \cite{bao2018backward}. An alternative approach employed a forward-backward stochastic differential equation (FBSDE) system without the doubly stochastic integral. This method approximates the filtering density at adaptively selected spatial points, using Shepard interpolation to reconstruct the density and MCMC resampling to mitigate degeneracy in the spatial point distribution \cite{bao2017adaptive}.

The kernel BSDE filter inherits the plain FBSDE and dynamics-driven adaptive spatial points from this independent line of work \cite{bao2017adaptive}. Instead of Shepard interpolation, the method employs kernel learning to construct a continuous approximation of the filtering density from the estimated density values at adaptive spatial points. Resampling is still performed, but drawing samples directly from the Gaussian mixture replaces MCMC-based resampling \cite{archibald2022kernel}.

Within this survey, the BSDE filter represents the filtering density as a kernel mixture fitted to density values computed at particle locations, and updates it by propagating those density values through a forward-backward stochastic differential equation system and then applying a Bayesian inference. This section adopts the mathematical formulations established in the kernel BSDE framework \cite{archibald2022kernel}.

Because the doubly stochastic integral term is no longer part of the backward equation, it reduces to the ordinary Fokker-Planck equation, the partial differential equation governing how the probability density function of a stochastic state evolves forward in time, rather than the Zakai equation the doubly stochastic branch relies on \cite{archibald2022kernel, zakai1969optimal}. Under appropriate regularity conditions, the Fokker-Planck density $p_t$ can be identified with the solution $Y_t$ of the backward equation in the Forward-Backward SDE (FBSDE) system \cite{pardoux2005backward}. This identification converts a high-dimensional PDE problem into a tractable Monte Carlo problem, since the FBSDE system can be sampled rather than solved on a grid.

The FBSDE system contains two coupled equations. The first propagates sample positions backward in physical time:
\begin{equation} \label{eq:bsde_state}
    X_{t_n} = X_{t_{n+1}} - \int^{t_{n+1}}_{t_n} b(X_s) \, ds 
    + \int^{t_{n+1}}_{t_n} \sigma_s \, d\overleftarrow{W}_s,
\end{equation}
where $\int \sigma_s \, d\overleftarrow{W}_s$ denotes the backward It\^{o} integral \cite{pardoux1994backward}. The second propagates density values forward:
\begin{equation} \label{eq:bsde_density}
    Y_{t_{n+1}}^{M_{t_n}} = Y_{t_n}^{M_{t_n}} 
    - \int^{t_{n+1}}_{t_n} \sum^d_{i=1} 
    \frac{\partial b_i}{\partial x_i}(X_s) Y_s^{M_{t_n}} \, ds
    - \int^{t_{n+1}}_{t_n} Z_s^{M_{t_n}} \, d\overleftarrow{W}_s,
\end{equation}
propagating from $Y_{t_n}^{M_{t_n}}$ to obtain the predicted filtering density $p(X_{t_{n+1}}|\mathcal{M}_{t_n})$ \cite{archibald2022kernel}. The term $Z_s^{M_{t_n}}$ is the martingale representation of $Y_s^{M_{t_n}}$ with respect to the Brownian motion $W$. It absorbs new randomness as the solution propagates backward, allowing $Y_s^{M_{t_n}}$ to remain a martingale at every point along the way rather than drifting away from the information available at that point \cite{FBSDE, BDSDE}. The two equations together implement the Chapman-Kolmogorov prediction step. The Bayesian update is applied separately once the predicted density values are available.

Because the FBSDE system cannot be solved analytically, the filter computes the density value at each spatial sample via a fixed-point iteration \cite{archibald2022kernel}:
\begin{equation} \label{eq:bsde_fixedpoint}
    Y_{n+1}^{M_{t_n}, l+1} = 
    \mathbb{E}_{n+1}^X[Y_n^{M_{t_n}}] 
    - \Delta t_n \sum_{i=1}^{d} 
    \frac{\partial b_i}{\partial x_i}(X_{t_{n+1}}) 
    Y_{n+1}^{M_{t_n}, l}, 
    \quad l = 0, 1, \ldots, L-1.
\end{equation}
Rather than computing the conditional expectation $\mathbb{E}_{n+1}^X[Y_n^{M_{t_n}}]$ as an expensive Monte Carlo average over many samples, the filter draws one new sample at each iteration and averages over the batch of samples accumulated across all iterations:
\begin{equation} \label{eq:bsde_rm}
    \tilde{\mathbb{E}}_{n+1}^{x,l}[Y_n^{M_{t_n}}] 
    = \frac{1}{l}\sum_{i=1}^{l} 
    Y_n^{M_{t_n}}(\tilde{X}_n^{x,i}),
\end{equation}
which converges to the true conditional expectation as iterations proceed \cite{archibald2022kernel}.

After computing predicted density values at all $N$ spatial samples, Bayesian inference incorporates the observational data \cite{archibald2022kernel}:
\begin{equation} \label{eq:bsde_update}
    \tilde{p}(X_{t_{n+1}} \mid \mathcal{M}_{t_{n+1}}) 
    = \frac{p(M_{t_{n+1}} \mid X_{t_{n+1}}) 
    \cdot Y_{n+1}^{M_{t_n}}}{\mathcal{Z}}.
\end{equation}
These updated density values serve as training targets for a Gaussian kernel mixture, fitted by stochastic gradient descent to produce a globally defined kernel density approximation over $\mathbb{R}^d$ \cite{archibald2022kernel}. Under Lorenz-96, the BSDE filter outperforms both the auxiliary particle filter and the Ensemble Kalman filter in terms of RMSE up to $d=20$ \cite{archibald2022kernel}.

The BSDE filter's representational ceiling is revealed in the paper's own scaling experiment: as dimension increases from $d=10$ to $d=20$, the number of spatial samples required grows from $800$ to $1{,}500$ and the number of kernels grows from $10$ to $20$, yet computation time still more than triples, from $5.71$ to $18.57$ seconds, for only a twofold increase in dimension \cite{archibald2022kernel}. In the three-property framework of this survey, the BSDE filter achieves non-Gaussian expressiveness and maintains scalability in terms of accuracy, but sacrifices affordable density representation, since the kernel mixture's computational cost grows faster than the resources allocated to maintain it.

Despite this representational limitation, the BSDE filter's most lasting contribution is that running a stochastic process backward in time extracts exactly the distributional information the filtering update requires. As the filtering problem scaled to high-dimensional chaotic systems with real operational consequences, this mathematical pursuit coalesced into data assimilation as a field, where the challenge is no longer only theoretical tractability but real-time performance at planetary scale. Score-based diffusion models \cite{song2020score}, developed independently in the machine learning community, provided the missing representational foundation: encoding the filtering density in an object whose cost does not grow with dimension or posterior complexity.

\section{Score-Based Filter and Ensemble Score Filter}
This section introduces two methods that realize the score-based approach within the filtering framework: the Score-based Filter (SF), which approximates the score via neural network training, and the Ensemble Score Filter (EnSF), which replaces the neural network with a closed-form training-free Monte Carlo estimator \cite{bao2024ensemble, bao2024score}. Both methods preserve the recursive Bayesian loop and realize the same backward stochastic mechanism the BSDE filter identified, now operating in pseudo-time rather than physical time, with the score function guiding the reverse process in place of the martingale correction term.

\subsection{Score-Based Filter (SF)}
In the terminology of this survey, the score-based filter represents the filtering density through its score function and updates it by running a reverse-time stochastic differential equation in pseudo-time guided by the posterior score. This section adopts the notation established in the score-based filtering framework \cite{bao2024score}.

The score function is defined as
\begin{equation} \label{eq:score_def}
    S(Z_\tau, \tau) := \nabla_z \log 
    Q_\tau(Z_\tau).
\end{equation}
The score is the gradient of the log-density at each point in the state space. It points toward regions of higher log-probability from any given location, equals zero at the modes, and encodes the local gradient structure of the log-density surface without storing density values directly \cite{song2020score}.

The score is defined over a pseudo-temporal domain $\tau \in [0,1]$, entirely separate from physical assimilation time. The BSDE system requires knowing the filtering density at $t_{n+1}$ as a terminal condition before the backward process can begin, but that density is precisely what the filter is trying to compute. In pseudo-time, the forward SDE
\begin{equation} \label{eq:fsde}
    dZ_\tau = b(\tau)Z_\tau \, d\tau + 
    \sigma(\tau)dW_\tau
\end{equation}
with drift and diffusion coefficients
\begin{equation} \label{eq:drift_diff}
    b(\tau) = \frac{d\log\alpha_\tau}{d\tau}, 
    \qquad
    \sigma^2(\tau) = \frac{d\beta_\tau^2}{d\tau} 
    - 2\,\frac{d\log\alpha_\tau}{d\tau}\,
    \beta_\tau^2
\end{equation}
maps any filtering density $Q(Z)$ at $\tau=0$ to $\mathcal{N}(0,I_d)$ at $\tau=1$, so the reverse process always starts from the same known reference point regardless of posterior shape, eliminating the terminal condition problem. The forward SDE is not used for sampling; it establishes that the conditional distribution of $Z_\tau$ given 
$Z_0$ is Gaussian:
\begin{equation} \label{eq:conditional}
    Q_\tau(Z_\tau | Z_0) = 
    \mathcal{N}(\alpha_\tau Z_0, 
    \beta_\tau^2 I_d),
\end{equation}
where $\alpha_\tau = 1 - \tau$ and $\beta_\tau^2 = \tau$, so at $\tau=1$ the marginal distribution is $\mathcal{N}(0,I_d)$.

The key result from the theory of reverse-time diffusions is that running a stochastic process backward in pseudo-time recovers samples from the target distribution without requiring an explicit parametric form \cite{anderson1982reverse}. The reverse-time SDE (RSDE)
\begin{equation} \label{eq:rsde}
    dZ_\tau = \left[b(\tau)Z_\tau 
    - \sigma^2(\tau)S(Z_\tau, 
    \tau)\right]d\tau 
    + \sigma(\tau)d\overleftarrow{W}_\tau
\end{equation}
maps standard Gaussian noise backward from $\tau=1$ to $\tau=0$, recovering samples from the filtering density. The score function $\mathbf{S}(Z_\tau, \tau)$ plays precisely the role the martingale term $Z_s^{M_{t_n}}$ played in the BSDE system: it keeps the backward evolution uniquely solvable and steers trajectories toward high-probability regions. The score function has a closed-form expression as a gradient of the log-density, computable directly at any point, whereas the martingale term has no closed form and must be approximated through a stochastic procedure at every spatial point at every step, which makes SF perform at higher dimensions than BSDE. Because the RSDE cannot be solved analytically, the Euler-Maruyama discretization is used \cite{kloeden1977numerical}:
\begin{equation} \label{eq:euler}
    Z_{\tau_k, j} = Z_{\tau_{k+1}, j} 
    - \left[b(\tau_{k+1}) Z_{\tau_{k+1}, j} 
    - \sigma^2(\tau_{k+1}) 
    \bar{S}_{t+1|t+1}
    (Z_{\tau_{k+1}, j}, \tau_{k+1})\right] 
    \Delta\tau
    + \sigma(\tau_{k+1}) \sqrt{\Delta\tau} 
    \, \xi_j,
\end{equation}
where $\xi_j \sim \mathcal{N}(0, I_d)$.

Rather than applying the Bayesian update as a single multiplicative step after density propagation, the score-based filter embeds it within the score function throughout the reverse process. Taking the gradient of the log-posterior via Bayes' theorem gives
\begin{equation} \label{eq:posterior_score_bayes}
    \nabla_x \log p(X_{t+1}|\mathcal{Y}_{t+1}) = 
    \nabla_x \log p(X_{t+1}|\mathcal{Y}_t) + 
    \nabla_x \log p(Y_{t+1}|X_{t+1}),
\end{equation}
so the posterior score decomposes as
\begin{equation} \label{eq:posterior_score}
    \bar{S}_{t+1|t+1}(Z_\tau, \tau) = 
    \bar{S}_{t+1|t}(Z_\tau, \tau) + 
    h(\tau)\nabla_z \log p(Y_{t+1}|Z_\tau),
\end{equation}
where $\bar{S}_{t+1|t}$ is the prior score estimated from the ensemble and
\begin{equation} \label{eq:damping}
    h(\tau) = 1-\tau, \quad h(0)=1, 
    \quad h(1)=0.
\end{equation}
The damping function $h(\tau)$ spreads the Bayesian update across all pseudo-time steps as a continuous guidance process: at each step, the likelihood gradient nudges the reverse trajectory by $h(\tau)\nabla_z \log p(Y_{t+1}|Z_\tau)$, growing from zero at $\tau=1$ to its full value at $\tau=0$ \cite{bao2024score}. No single normalization step occurs, and every trajectory is guided independently rather than competing for weight, avoiding weight collapse under strongly nonlinear observations. The likelihood gradient is computed analytically from the observation model, requiring no spatial discretization \cite{bao2024ensemble, bao2024score}.

On the Lorenz-96 system, the score-based filter achieves competitive accuracy against the online variational filter at $d=100$ \cite{bao2024score}, extending the practical range by a factor of five compared to the BSDE filter's ceiling of $d=20$. However, two constraints limit its operational feasibility. First, approximating the score function requires training a neural network at each assimilation step, demanding substantial wall-clock time and making real-time deployment infeasible \cite{bao2024ensemble}. Second, training requires storing all paths of the forward SDE, imposing high memory cost. These constraints motivate a training-free score approximation, which is the central contribution of the Ensemble Score Filter introduced in the following subsection.

\subsection{Ensemble Score Filter (EnSF)}
The score-based filter establishes that encoding the filtering density in the score function avoids both the kernel mixture scalability ceiling and the one-shot Bayesian weight collapse. Its remaining constraint is that the score function requires neural network retraining at every assimilation step. The Ensemble Score Filter (EnSF) resolves this by reformulating the score estimation as a closed-form Monte Carlo problem, making the score approximation training-free at every assimilation step \cite{bao2024ensemble}. In the terminology of this survey, EnSF represents the filtering density through its score function and updates it by running the reverse-time SDE guided by a training-free importance-weighted Monte Carlo score estimator.

The training-free score estimation exploits the Gaussian conditional structure of the linear forward SDE, which yields a closed-form expression for the score as an importance-weighted expectation over the prior ensemble. This principle has been established in recent literature, where theoretical consistency and convergence of the Monte Carlo estimator are proven under Lipschitz continuity of the score and square-integrability of the ensemble \cite{wang2026error}. EnSF derives a closed-form expression for the prior score from the transition probability of the FSDE:
\begin{equation} \label{eq:ensf_score}
    S_{t|t-1}(z_{t,\tau}, \tau) = \int_{\mathbb{R}^d} 
    -\frac{z_{t,\tau} - \bar{\alpha}_\tau z_{t,0}}
    {\bar{\beta}_\tau^2} \, 
    w(z_{t,\tau}, z_{t,0}) \, 
    q_{Z_{t,0}}(z_{t,0}) \, dz_{t,0},
\end{equation}
where $w(z_{t,\tau}, z_{t,0})$ is the normalized importance weight
\begin{equation} \label{eq:ensf_weight}
    w(z_{t,\tau}, z_{t,0}) = 
    \frac{q_{Z_{t,\tau}|Z_{t,0}}(z_{t,\tau} | z_{t,0})}
    {\int q_{Z_{t,\tau}|Z_{t,0}}(z_{t,\tau} | z'_{t,0}) 
    q_{Z_{t,0}}(z'_{t,0}) dz'_{t,0}}.
\end{equation}

EnSF introduces regularized variants of the signal and noise scaling parameters $\alpha_\tau$ and $\beta_\tau$ defined in the score-based filter. In the score function, the denominator $\bar{\beta}_\tau^2 \rightarrow 0$ as $\tau \rightarrow 0$, causing the score estimate to diverge. Similarly, $\alpha_\tau = 1-\tau \rightarrow 0$ removes all signal from the ensemble near $\tau = 0$, collapsing the prior information. EnSF addresses both by introducing small regularization constants $\epsilon_\alpha$ and $\epsilon_\beta$:
\begin{equation} \label{eq:ensf_reparam}
    \bar{\alpha}_\tau = 1 - \tau(1 - \epsilon_\alpha), 
    \qquad 
    \bar{\beta}_\tau^2 = \epsilon_\beta + 
    \tau(1 - \epsilon_\beta),
\end{equation}
ensuring $\bar{\alpha}_\tau \geq \epsilon_\alpha > 0$ and $\bar{\beta}^2_\tau \geq \epsilon_\beta > 0$ across the full interval $[0,1]$, keeping the score estimate finite and the reverse-time SDE stable.

This expression still contains an integral over $q_{Z_{t,0}}$ that cannot be evaluated exactly, since the ensemble's distribution is not known in closed form. EnSF approximates it via a Monte Carlo mini-batch sum over the predicted ensemble:
\begin{equation} \label{eq:ensf_mc}
    \bar{S}_{t|t-1}(z_{t,\tau}, \tau) \approx 
    \sum_{n=1}^{N} 
    -\frac{z_{t,\tau} - \bar{\alpha}_\tau 
    f(x_{t-1,j_n}, \omega_{t-1,j_n})}{\bar{\beta}_\tau^2} 
    \cdot \bar{w}(z_{t,\tau}, 
    f(x_{t-1,j_n}, \omega_{t-1,j_n})).
\end{equation}
In practice, the normalized weights $\bar{w}$ concentrate nearly all mass on a single ensemble member, making additional terms in the sum redundant \cite{wang2026error}. The batch size is therefore set to one, justified both by the convergence analysis establishing theoretical consistency and by empirical confirmation across all tested configurations \cite{bao2024ensemble, wang2026error}.

Weight concentration has a different meaning depending on the method used. In the particle filter, weights attach permanently to a fixed set of particles that is the entire posterior representation. If the weights collapse onto one particle, the filter has lost the posterior everywhere except wherever that particle is located. In EnSF, the weight $\bar{w}$ in Eq.~\ref{eq:ensf_mc} is instead recomputed fresh at every query point $z_{t,\tau}$ visited during the reverse diffusion. The object being estimated is a local score, not a global density the entire ensemble must jointly sustain, which is why concentrated weights justify batch size one for EnSF without reproducing the particle filter's degeneracy.

\begin{algorithm}[H]
\caption{Ensemble Score Filter (EnSF)}
\begin{algorithmic}[1]
\State \textbf{Input:} state model $f(X_t, \omega_t)$, 
prior density $P(X_0)$
\State Generate $J$ samples $\{x_{0,j}\}_{j=1}^{J}$ 
from $P(X_0)$
\For{$t = 1, 2, \ldots$}
    \State Propagate ensemble: 
    $\{f(x_{t,j}, \omega_{t,j})\}_{j=1}^{J}$ 
    via state equation
    \For{$k = K-1, \ldots, 0$}
        \State Compute importance weights $\bar{w}$ 
        via Eq.~\ref{eq:ensf_weight}
        \State Compute prior score 
        $\{\bar{S}_{t+1|t}(z, \tau)\}_{j=1}^{J}$ 
        via Eq.~\ref{eq:ensf_mc}
        \State Compute posterior score 
        $\{\bar{S}_{t+1|t+1}(z,\tau)\}_{j=1}^{J}$ 
        via Eq.~\ref{eq:posterior_score} (Section~4.1)
        \State Update ensemble 
        $\{z_{\tau_k}\}_{j=1}^{J}$ 
        via Eq.~\ref{eq:euler} (Section~4.1)
    \EndFor
    \State Set $\{x_{t+1,j}\}_{j=1}^{J} = 
    \{z_{0,j}\}_{j=1}^{J}$
\EndFor
\end{algorithmic}
\end{algorithm}

EnSF simultaneously escapes all three classical constraints. The score function representation places no parametric constraint on the posterior shape, giving EnSF non-Gaussian expressiveness \cite{bao2024ensemble, bao2024score}. The training-free Monte Carlo estimator has been demonstrated on systems with state dimensions exceeding one million \cite{bao2024ensemble}, establishing scalability. By encoding the filtering density in the score function rather than in discrete samples or kernel mixtures, the representational cost does not grow with dimension or posterior complexity. These three properties establish EnSF as the foundation for the comparison this survey runs next: whether they hold up not just in principle, but against EnKF, the particle filter, and LETKF on a chaotic, high-dimensional benchmark.

\subsection{Experiment}
This section conducts two experiments: the first between EnSF, EnKF, and bootstrap PF, and the second between EnSF and LETKF. All code is publicly available at \url{https://github.com/cathpham/lorenz96-da-benchmark}.

\subsubsection{EnSF vs. EnKF and PF}

The first experiment heavily tests whether each filter can track the true trajectory under a nonlinear observation operator as the dimension increases. The experiment operates on Lorenz-96 system, defined by
\begin{equation} \label{eq:lorenz96}
    \frac{dX_i}{dt} = (X_{i+1} - X_{i-2})\,X_{i-1} 
    - X_i + F, \qquad i = 1,\ldots,d,
\end{equation}
with forcing constant $F = 8$, which produces strongly chaotic dynamics \cite{lorenz1996predictability}. The truth is initialized on the attractor by drawing from $\mathcal{N}(0, 3^2 I_d)$ and integrating for 1000 fourth-order Runge-Kutta (RK4) steps. The true trajectory is then integrated deterministically with RK4. The forecast model is $dX = f(X)dt + \sigma_{\text{dyn}}dW_t$, with $f$ from \eqref{eq:lorenz96}, and discretized using the Euler-Maruyama scheme with $\sigma_{\text{dyn}}=0.2$. Both use the time step $\Delta t = 0.01$, and all states are clipped to $[-50,50]$ after each integration step for numerical stability. The observations is $g(X) = \arctan(X)$ and corrupted by Gaussian noise $\varepsilon \sim \mathcal{N}(0, \sigma_{\text{obs}}^2 I)$ with $\sigma_{\text{obs}} = 0.1$, with all $d$ components observed. Observation takes place every 10 model steps over $T=1500$ steps, making 150 assimilation cycles. The ensemble initialization is centered at the true initial state $x_{0,j} \sim \mathcal{N}(X_0, 0.1^2 I_d)$ for $j = 1,\ldots,N$.

All methods are tested with dimension $d=5,10,50$. The ensemble size is $N = 20$ for EnKF and EnSF, while the particle size for PF is $10{,}000$, due to the fact that PF heavily depends on particles to assimilate data. EnKF is the stochastic (perturbed-observation) variant without inflation or localization. EnSF solves the RSDE with 100 Euler steps per assimilation cycle. Its schedule hyperparameters are tuned for each tested dimension: for $d=5$ $\epsilon_\alpha = 0.75$ and $\epsilon_\beta = 0.025$, for $d=10$ $\epsilon_\alpha = 0.65$ and $\epsilon_\beta = 0.05$, for $d=50$ $\epsilon_\alpha = 0.55$ and $\epsilon_\beta = 0.0625$. For PF, it performs systematic resampling whenever the ESS drops below $N/2$, followed by the Gaussian kernel jittering (regularization) of duplicated particles, scaled by the weighted sample covariance. The RMSE is the error of the ensemble mean (weighted mean for PF) against the truth, averaged over 10 seeds.

\begin{figure}[H]
    \centering
    \includegraphics[width=1\linewidth]{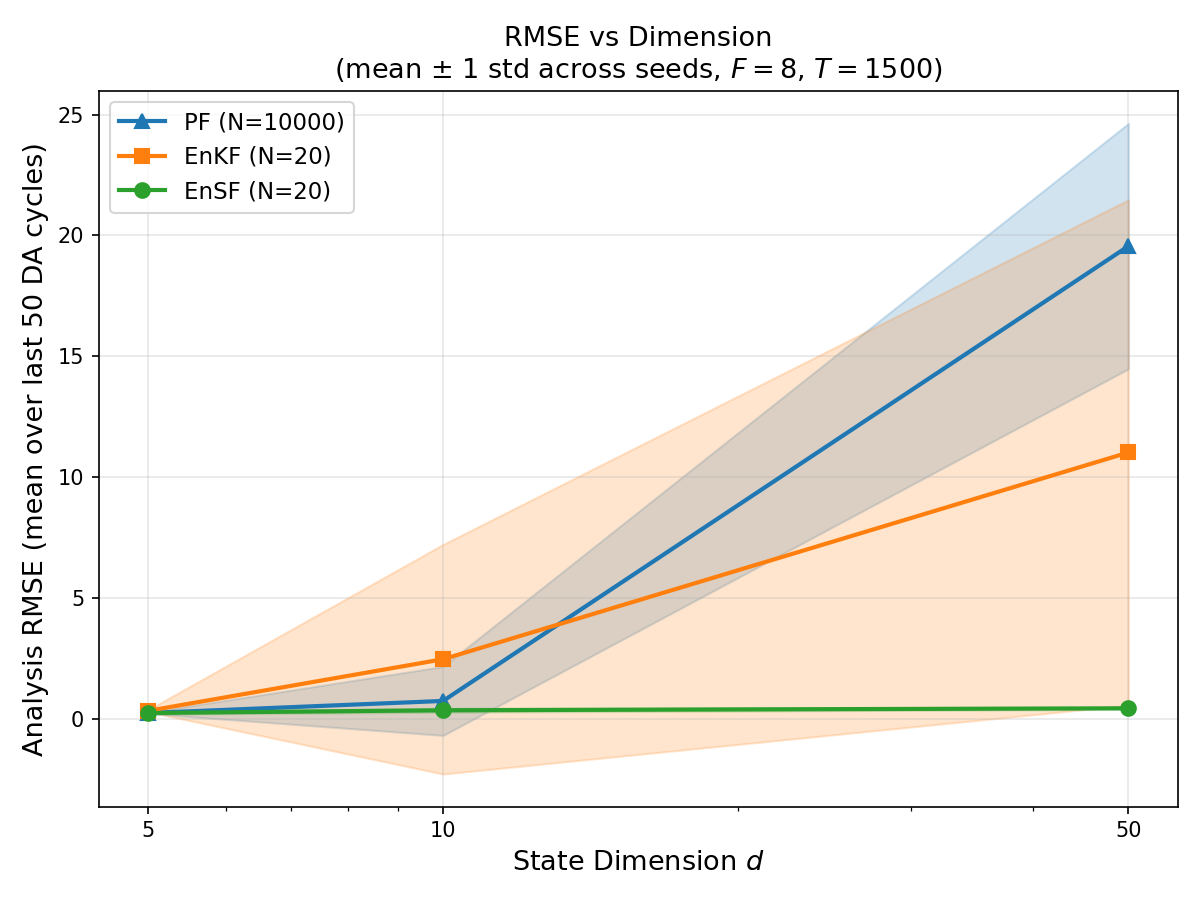}
    \caption{Mean RMSE over the last 50 assimilation cycles as a function of state dimension $d$, averaged over 10 independent runs. Shaded bands represent one standard deviation across runs.}
    \label{fig:rmse_vs_dim}
\end{figure}

EnKF performs well at $d=5$, as its RMSE is approximately $0.35$. However, EnKF's RMSE rises sharply to $2.46$ at $d = 10$ and $11.02$ at $d = 50$, along with growing spread across seeds. It is because EnKF's linear updates cannot represent the non-Gaussian posterior (arctangent imposes strong nonlinearity), which creates analysis bias \cite{carrassi2018data}. This error accelerates as the dimension grows, since a fixed ensemble size $N=20$ makes the sample covariance rank-deficient at $d=50$, generating spurious correlations between distant state components \cite{carrassi2018data}.

PF's results are accurate at $d=5$ (RMSE $\approx$ $0.24$) and remain reasonably accurate at $d=10$ (RMSE $\approx$ $0.74$). In low dimensions, $N=10{,}000$ particles is sufficient to cover the state space and represent the non-Gaussian filtering density well. However, at $d=50$, PF diverges to an RMSE of around $19.56$, with larger variability across different runs. Even with resampling and kernel jittering, the weights collapse onto a few particles and PF loses track of the true trajectory. Larger dimensions create a larger state space and a more concentrated likelihood \cite{snyder2008obstacles}. The same number of particles cannot keep up with the exponentially growing space as the dimension grows, resulting in the curse of dimensionality \cite{snyder2008obstacles}.

EnSF maintains a low RMSE of approximately $0.23$, $0.36$, and $0.44$ across all three tested dimensions with a narrow uncertainty band throughout. At $d=5$, its RMSE is comparable to that of PF. Unlike EnKF, EnSF does not rely on a linear-Gaussian update, as its likelihood gradient $\nabla_z \log p(Y_{t+1}|Z_\tau)$ is computed analytically from the arctangent observation model, so it can represent non-Gaussian posterior \cite{bao2024ensemble}. EnSF also does not use importance weights like PF, so it avoids weight collapse as the dimension grows \cite{bao2024ensemble}. As a result, EnSF remains accurate at $d=50$ with the same $N=20$ members as EnKF. 

\begin{figure}[H]
    \centering
    \includegraphics[width=1\linewidth]{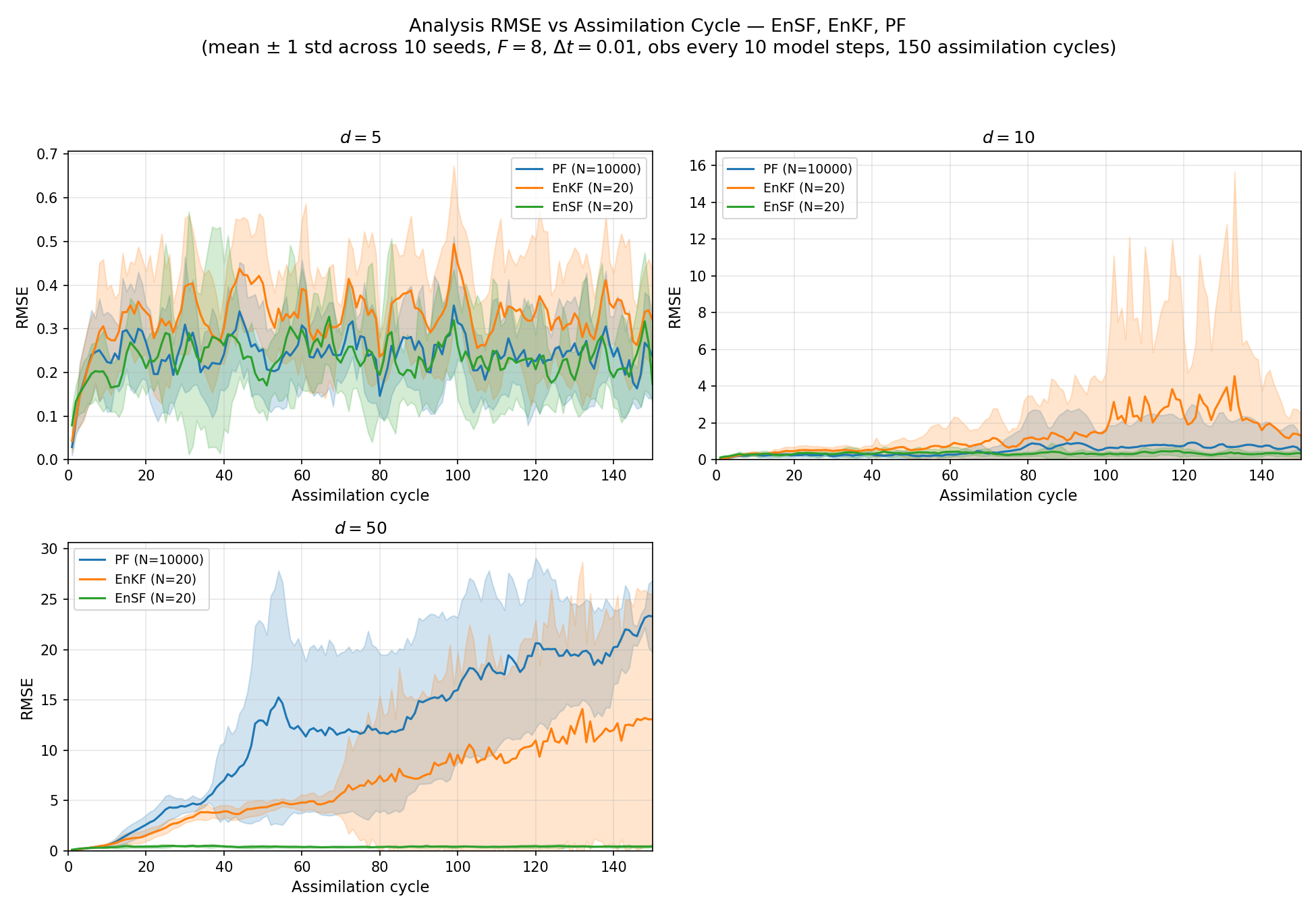}
    \caption{RMSE at each filtering step across four dimensions ($d=5, 10, 50$), averaged over 10 independent runs. Shaded bands represent one standard deviation across runs.}
    \label{fig:rmse_vs_step}
\end{figure}

Since all ensembles are initialized around the true state, their RMSE starts near zero and rises to a steady level within the first 10 cycles. At $d=5$, EnKF tracks the truth throughout the run, and its RMSE fluctuates around $0.35$. Yet, this error is generally above the ones from PF and EnSF. At a low dimension, with $N=20 \gg d$, EnKF estimates the covariance accurately, and its error comes from a small bias of the linear update under nonlinear observation operator. EnSF and PF achieve better results because both can represent non-Gaussian structure.

At $d=10$, EnKF starts to degrade after about assimilation cycle 70, and after assimilation cycle 100, its RMSE fluctuates between 2 and 4 with several spikes, with larger standard deviation. Since $N=20 > d$, EnKF's covariance is still full-rank, so this degradation mainly comes from its bias of the linear update under arctangent observation. At a similar cycle, PF also shows larger variability, yet its estimates remain fairly accurate. EnSF maintains a low RMSE throughout 150 assimilation cycles.

At $d=50$, both EnKF and PF start to lose track of the true trajectory at around assimilation cycle 10. PF degrades faster, as its variability gets larger at around cycle 35, and its RMSE reaches around 23 by the end of the run, compared with cycle 70 and about 13 for EnKF. At this dimension $N-1=19 < d$, resulting in EnKF's rank-deficiency, which amplifies the bias of its linear update. PF suffers from the curse of dimensionality as its $N=10{,}000$ is no longer enough to cover the state space. EnSF maintains stable and low RMSE throughout 150 assimilation cycles, with very narrow spread across seeds and no sign of divergence.

\begin{figure}[H]
    \centering
    \includegraphics[width=1\linewidth]{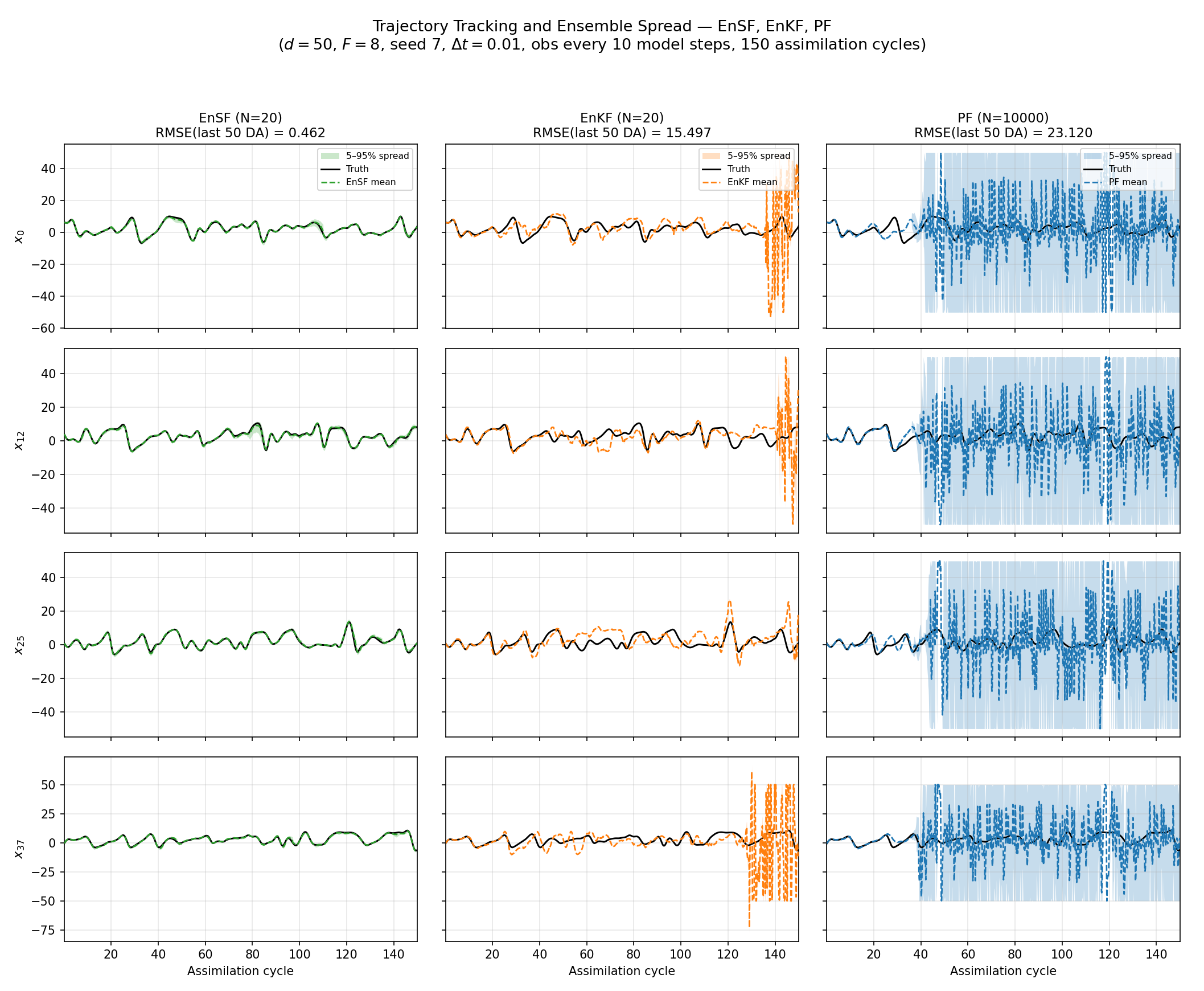}
    \caption{Trajectory tracking and ensemble spread for four representative state components ($x_0, x_{12}, x_{25}, x_{37}$) at $d=50$. Each column shows one method's posterior mean against the true state alongside the 5th to 95th percentile ensemble spread.}
    \label{fig:trajectory}
\end{figure}

EnKF follows the shape of truth until around assimilation cycle 30, but then its mean gradually departs from the true trajectory at several peaks and troughs. From around cycle 130 to 150, depending on the component, EnKF loses track completely, oscillating violently and reaching values of $\pm 50$ or beyond on $x_0$, $x_{12}$, and $x_{37}$. Meanwhile, EnKF's ensemble spread remainsnarrow relative to its error, illustrating its overconfidence in biased estimates.

PF tracks the system up to around assimilation cycle 40, then loses track of all four components almost simultaneously. From that point on, its mean fluctuates erratically with no relation to the true state, and its ensemble spread fills nearly the entire clipped range. Hence, PF's estimation carries no useful information about the truth.

EnSF tracks all four components accurately throughout all 150 assimilation cycles. Its posterior mean follows every peak and trough of the chaotic trajectory, and its ensemble spread remains narrow and consistent with its small error. Together, the three columns show that, in this run, EnSF is the only method that maintains accurate tracking of the truth across the state components over the full assimilation window.

\subsubsection{EnSF vs. LETKF}

The second experiment compares EnSF and the current state-of-the-art LETKF in a high dimensional and partially observed setting. It shares with first experiment model, truth initialization, time step, clipping bound, observation operator, schedule, and number of seeds. However, there are several differences. Both the true trajectory and the forecast model are integrated deterministically with RK4, which makes the forecast model perfect. The observations only take place every second component, fixed in every assimilation cycle and in every seed, with $\sigma_\text{obs}=0.05$. The ensemble is initialized with no knowledge of the true state, $\mathcal{N}(0,I_d)$.

EnSF and LETKF are tested at $d = 1000$, with ensemble size $N = 100$. EnSF solves the RSDE with 100 Euler steps per assimilation cycle, with $\epsilon_\alpha = 0.525$, $\epsilon_\beta = 0.045$. For LETKF, the inflation is $0.915$ and the localization radius is $1.0$. The RMSE is the error of the ensemble mean against the truth, computed over all components and separately over the observed and unobserved components, and averaged over 10 seeds. Both methods run on a GPU, and the wall-clock time of each assimilation step is recorded.

\begin{figure}[H]
    \centering
    \includegraphics[width=1.0\linewidth]{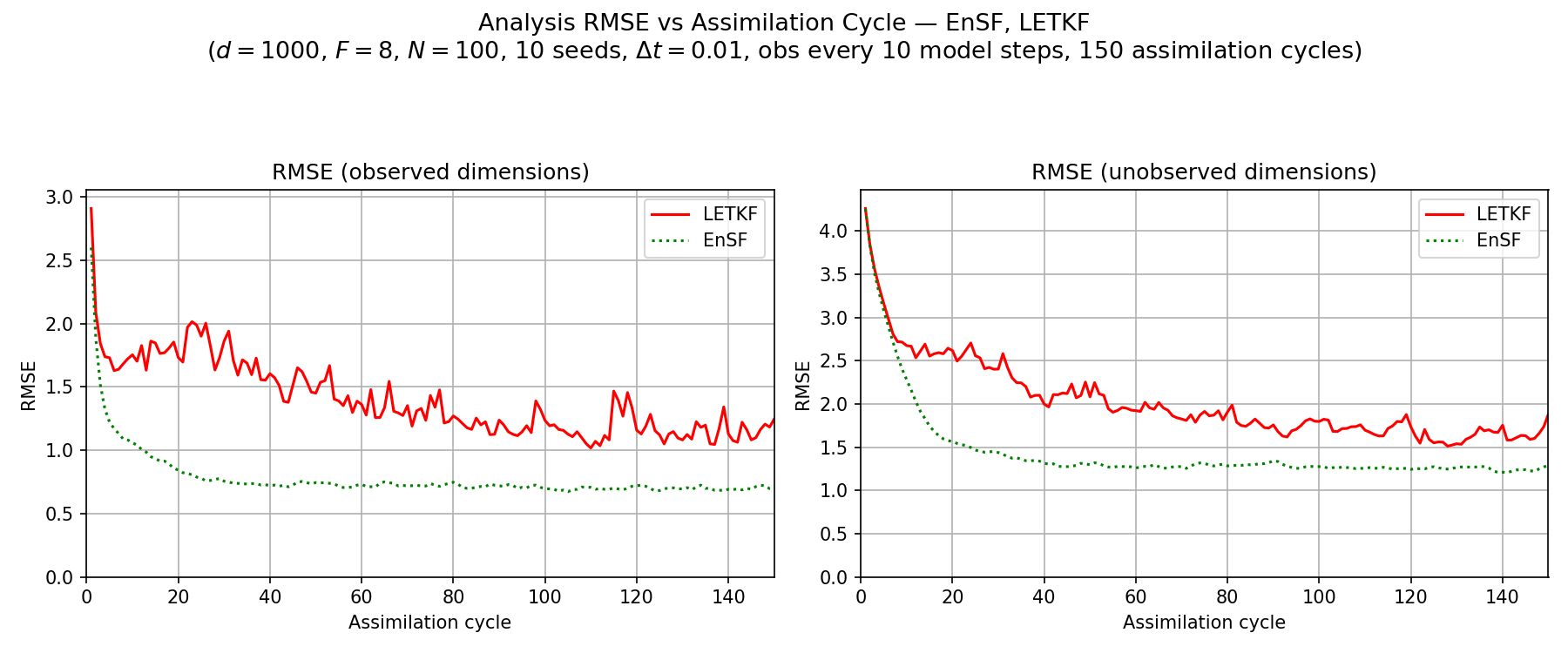}
    \caption{LETKF versus EnSF: RMSE across model time for observed and unobserved dimensions.}
    \label{fig:ensf_vs_letkf}
\end{figure}

Figure \ref{fig:ensf_vs_letkf} compares EnSF against LETKF across both observed and unobserved dimensions. Two hyperparameters tuned separately for each tested dimension, localization radius and multiplicative inflation factor. Both methods recover the state from an uninformed start, making their RMSE starts high ($\approx 2.6-2.9$ for observed dimensions, and $\approx 4.3$ unobserved), but then both curves drop quickly.

EnSF converges quickly, as its RMSE drops to approximately $0.7$ by assimilation cycle 30 for the observed components, and to $1.3$ by cycle 40 for the unobserved components. Its curve remains flat at these low values. In contrast, LETKF takes around 40 cycles for RMSE to reach $1.5$ for observed dimensions, and then gradually drop to $1.2$ by the end of the run. For unobserved dimensions, LETKF stays near $2.5$ and only gradually decreases after cycle 25. Its curve is also noisier than that of EnSF throughout the run.

The computational cost panel reverses the comparison. By $d=1000$, LETKF requires approximately $1.8$ seconds per assimilation step against EnSF's $0.05$ seconds, based on experiments run on an Google Colab GPU T4. At $d=1{,}000{,}000$, EnSF completes each step in approximately $5$ seconds while LETKF requires $300$ seconds, a $60\times$ difference reflecting the architectural constraint of localization-based methods. These large-scale timing results are reproduced from the original EnSF paper, conducted on an AMD Ryzen 5 5600X CPU and an NVIDIA RTX 3070 GPU \cite{bao2024ensemble}. EnSF's computational structure fit naturally onto GPU memory, and EnSF with GPU implementation is substantially more efficient than LETKF and more suitable for the GPU-based supercomputers that now dominate high-performance scientific computing \cite{bao2024ensemble}.

Across both experiments, EnSF is the only method that achieves non-Gaussian posterior, low and stable accuracy across the tested dimensions, from $d=5$ to $d=1000$, with lower computational cost. These four properties suggest that EnSF is a promising foundation for high-dimensional nonlinear filtering. However, the experiments above are limited to Lorenz-96, which has spatially uniform dynamics, idealized model error, and regular observation networks. The following section tests whether this advantage holds beyond a synthetic benchmark, on a higher-dimensional system with realistic geophysical turbulence.

\subsection{Validation Beyond Synthetic Benchmarks}
Beyond the synthetic Lorenz-96 benchmarks in Section 4.3, EnSF has also been applied to the surface quasi-geostrophic (SQG) model, a chaotic system with realistic geophysical turbulence, and a state dimension of 8192 \cite{bao2025nonlinear}. Compared against the operational LETKF across four experimental settings of increasing difficulty, EnSF matches or trails LETKF only in the fully linear, perfect-model case, where Kalman-filter theory expects LETKF to be optimal \cite{bao2025nonlinear}. Once the model is corrupted by unpredictable errors, LETKF's analysis errors spike to up to four times the average EnSF error even with near-optimally tuned localization and inflation \cite{bao2025nonlinear}. Under both nonlinear arctangent observations, EnSF systematically outperforms the best-tuned LETKF configuration while using no localization or inflation tuning at all \cite{bao2025nonlinear}. This confirms, in a higher-dimensional and physically realistic chaotic system, the same pattern established on synthetic benchmarks: EnSF's advantage over LETKF grows precisely as the assimilation problem departs from the linear-Gaussian assumptions LETKF depends on.

A direct extension of this SQG study examines whether EnSF's robustness holds not just across experimental settings but across the design of the observation network itself, varying both the spatial distribution of observations and the fraction that are nonlinear \cite{xiong2026robustness}.  With fully linear observations, LETKF remains roughly four times more accurate than EnSF across all three networks (RMSE $\approx 0.6$--$0.7$ versus $\approx 2.4$--$2.8$), but once any nonlinear observations are introduced, LETKF's error jumps to $\approx 8$--$10$ and stays there regardless of network type, while EnSF's RMSE remains flat at $\approx 2.4$--$2.8$ across every network and every nonlinear fraction tested \cite{xiong2026robustness}. A parallel tuning experiment under observation locations change every analysis  finds that LETKF's viable range of localization and inflation parameters collapses as nonlinearity increases, leaving only one stable configuration at 5\% nonlinear observations in the network with locations resampled at every analysis step, and none at 15\%, while EnSF requires no such tuning to remain stable \cite{xiong2026robustness}. This shows that EnSF's advantage under nonlinear observations is not an artifact of one fixed observation layout, strengthening the pattern established earlier in this section that EnSF's robustness generalizes across problem settings.

\section{Recent Advances and Enhancements}
The validation above shows EnSF's mechanism holds up on a real geophysical system, without any modification. The fourteen extensions below test how far that same mechanism stretches across new and harder problems, and where it finally has to change.

Act 0 improves the algorithm itself, before it is tested against anything external. IEnSF \cite{IEnSF} eliminates structural bias in the posterior score estimator, and EnFF \cite{EnFF} replaces the reverse-time SDE with a more efficient ODE formulation.

Act I then asks whether the resulting mechanism needs to change at all when pointed at harder problems, and finds that it does not. Two-Phase Flow EnSF \cite{hu2025ensemble} and Adaptive Learning EnSF for SPDEs \cite{huynh2025scorebaseddiffusionmodelapproach} generalize what produces the state, from one hardcoded solver to any SPDE solver. Fire-EnSF \cite{Fire-EnSF} generalizes the state's own size, handling a dimension that grows as a wildfire's perimeter expands. United Filter \cite{UF} generalizes what is being estimated, extending to joint state-parameter estimation. None of the four touch the score update itself.

Act II tests the mechanism against a different kind of forward model entirely, a machine learning forecaster whose reliability only extends to inputs it was trained on. Score-DA \cite{tang2026score} and EnSF for Real-Data Energy Consumption \cite{hu2026diffusion} correct the resulting drift without altering a single equation of the filter itself.

Act III marks the point where this restraint finally breaks under external demand. EnSF-LR \cite{xiong2026twostepensemblescorefilter}, Inpainting-EnSF \cite{Inpainting}, Latent-EnSF \cite{Latent-EnSF}, and LD-EnSF \cite{LD-EnSF} each pay an escalating price to correct components the observations never touch, a demand the original mechanism cannot meet on its own.

Act IV is this claim's honest limit. Until now, every method's own filtering step, whatever it depended on upstream, has remained training-free. DBUQ \cite{lu2024diffusion} and STORM \cite{wang2026global} finally surrender that property, first partially and then completely, in exchange for amortized Earth-system calibration and planetary-scale assimilation on exascale hardware.

Together, these five acts make a claim no single extension could: the score-based filtering framework is not fourteen unrelated fixes, but one mechanism tested against escalating operational demands. They absorb most of them unchanged, require real modification only when observations leave part of the state untouched, and give up its defining training-free property only at planetary scale. 

This survey reflects the state of the Ensemble Score Filter and its extensions up to the time of writing. As the field continues to develop, new methods and applications are expected to emerge and reshape existing results, including assimilation of aggregated surveillance data into agent-based epidemic models \cite{liang2026score}, covariance-informed score transport for partially observed states \cite{liang2026covariance}, the sensitivity of ensemble filters to the structure of the observation network \cite{xiong2025sensitivity}, and amortized conditional inference for uncertainty quantification \cite{zhang2025exact}.

\begin{table}[H]
\centering
\small
\caption{Capability comparison of filtering methods, scored only on what each paper experimentally demonstrated, not on what its mechanism could plausibly support. \checkmark\ indicates the property was explicitly tested and confirmed; $\sim$ indicates partial or condition-dependent satisfaction, details down below; an empty cell indicates the property was not demonstrated, even if the underlying mechanism may in principle allow it.}
\label{tab:capability}
\setlength{\tabcolsep}{4pt}
\begin{tabular}{lccccccc}
\hline
\textbf{Method} & 
\textbf{Non-linear} & 
\textbf{$\ge$1000D} & 
\textbf{Training-free} & 
\textbf{Sparse obs.} & 
\textbf{Var. topology} & 
\textbf{Joint est.}\\
\hline
KF                 &              &              & \checkmark &              &              &                           \\
EKF                & \checkmark   &              & \checkmark &              &              &                           \\
UKF                & \checkmark   &              & \checkmark &              &              &                           \\
EnKF               & \checkmark   &              & \checkmark & \checkmark   &              &                           \\
LETKF              & \checkmark   & \checkmark   & \checkmark & \checkmark   &              &                           \\
PF                 & \checkmark   &              & \checkmark &              &              &                           \\
APF                & \checkmark   &              & \checkmark &              &              & \checkmark        \\
RBPF               & \checkmark   &              & \checkmark & \checkmark   &              &                           \\
RPF                & \checkmark   &              & \checkmark &              &              &                           \\
BSDE filter        & \checkmark   &              &            &              &              &                           \\
SF                 & \checkmark   &              &            &              &              &                           \\
EnSF               & \checkmark   & \checkmark   & \checkmark &              &              &                           \\
IEnSF              & \checkmark   & \checkmark   & \checkmark & \checkmark   &              &                           \\
EnFF               & \checkmark   & \checkmark   & \checkmark &              &              &                           \\
Two-Phase Flow     & \checkmark   & \checkmark   & \checkmark & \checkmark   &              &                           \\
SPDE Adpt. Lrn     & \checkmark   & \checkmark   & \checkmark & \checkmark   &              &                           \\
Fire-EnSF          & \checkmark   & \checkmark   & \checkmark &              & \checkmark   &                           \\
United Filter      & \checkmark   & \checkmark   & \checkmark & \checkmark   &              &  \checkmark        \\
Score-DA           & \checkmark   &              & $\sim$     & \checkmark   &              &                           \\
Forecast Corrt.    & \checkmark   & \checkmark   & $\sim$     & \checkmark   &              &                           \\
EnSF-LR            & \checkmark   &              & \checkmark & \checkmark   &              &                           \\
Inpainting-EnSF    & \checkmark   & \checkmark   & \checkmark & \checkmark   &              &                           \\
Latent-EnSF        & \checkmark   & \checkmark   & $\sim$     & \checkmark   &              &                           \\
LD-EnSF            & \checkmark   & \checkmark   & $\sim$     & \checkmark   &              &                           \\
DBUQ               & \checkmark   &              &            &              &              &                           \\
STORM              & \checkmark   & \checkmark   &            & \checkmark   &              &                           \\
\hline
\end{tabular}
\end{table}

Training-free-ness does not transfer uniformly across the EnSF extensions surveyed. In Latent-EnSF, LD-EnSF, Score-DA, and Energy-Consumption Forecast Correction, the core assimilation step executed at deployment remains EnSF's original closed-form, training-free update \cite{Latent-EnSF, LD-EnSF, tang2026score, hu2026diffusion}, but this step is gated behind a component that must itself be trained: a coupled VAE in Latent-EnSF \cite{Latent-EnSF}, a Latent Dynamics Network and LSTM encoder in LD-EnSF \cite{LD-EnSF}, an LSTM/R-DeepONet forecast surrogate in Score-DA \cite{tang2026score}, and a pretrained spatio-temporal transformer in Energy-Consumption Forecast Correction \cite{hu2026diffusion}. Therefore these four are marked as partially training-free. This differs from DBUQ and STORM, both trained without qualification. DBUQ's score estimation is itself training-free \cite{lu2024diffusion}, but serves only to generate offline labeled data for a supervised generator network; at inference the reverse-time SDE is never solved, so no training-free computation survives to deployment. STORM contains no training-free step at any stage, defining its prior score directly as an end-to-end trained denoiser \cite{wang2026global}. The distinction, then, is not whether training occurs in the pipeline, but whether a training-free computation is the operational core actually run at deployment.

\subsection{Iterative EnSF (IEnSF)}
EnSF's posterior score combines the prior score and the likelihood gradient through a heuristic damping function $h(\tau) = 1 - \tau$ that enforces consistency with the true posterior score only at the endpoints $\tau = 0$ and $\tau = 1$ \cite{IEnSF}. At every intermediate pseudo-time where the reverse-time SDE actually performs its sampling, the posterior score carries uncorrected structural bias \cite{IEnSF}. Furthermore, when the observation operator covers only a subset of the state, the likelihood gradient vanishes along unobserved dimensions entirely, leaving those components unchanged by the Bayesian update regardless of their physical correlation with observed variables \cite{IEnSF}.

The key insight is that the true posterior score has an exact closed-form expression where the prior is modeled as a Gaussian mixture, and this exact expression is a covariance-aware structure that connects observed and unobserved dimensions through ensemble correlations \cite{IEnSF}. IEnSF preserves EnSF's core score-based representation and RSDE sampling mechanism while replacing the heuristic scalar damping with this analytically derived covariance-aware posterior score, refined iteratively using posterior samples from each RSDE solve \cite{IEnSF}.

Under a Gaussian mixture prior with $K$ components, the exact posterior score takes the form
\begin{equation} \label{eq:iensf_score}
    S_{t+1 \mid t+1}(Z_\tau \mid y_{t+1}) = 
    \sum_{k=1}^{K} \bar{w}^{\text{post}}
    (k, Z_\tau, y_{t+1})\, S_{Z_\tau|k}(Z_\tau) 
    + J(\tau)\, \mathbb{E}\big[\nabla_z \log 
    p(y_{t+1} \mid Z_0) \,\big|\, Z_\tau, 
    y_{t+1}\big],
\end{equation}
where $\bar{w}^{\text{post}}(k, Z_\tau, y_{t+1})$ is the posterior mixture weight of the $k$-th Gaussian component given the current reverse trajectory point $Z_\tau$ and observation $y_{t+1}$, and $S_{Z_\tau|k}(Z_\tau)$ is the score of the $k$-th component evaluated at $Z_\tau$ \cite{IEnSF}. The time-dependent matrix $J(\tau) = \bar{\alpha}_\tau \Sigma_{\text{ens}} (\bar{\alpha}_\tau^2 \Sigma_{\text{ens}} + \bar{\beta}_\tau^2 I_d)^{-1}$ replaces the scalar $h(\tau)$, incorporating the ensemble covariance $\Sigma_{\text{ens}}$ to eliminate the need for an ad hoc guidance strength, with the scaling adapting automatically to the ensemble covariance structure \cite{IEnSF}. Because this expectation is generally intractable, IEnSF instead evaluates the likelihood gradient at a single reference point $\bar{\mu}^*_0(Z_\tau)$, the conditional mean EnSF fixed heuristically \cite{IEnSF}. Because the true posterior mean is unknown, IEnSF refines this reference point iteratively: each RSDE solve produces posterior samples that update $\bar{\mu}^*_0$, which drives the posterior mixture weights $\bar{w}^{\text{post}}$ in the next iteration until convergence \cite{IEnSF}.

In 2D synthetic experiments spanning a linear consistency check and two nonlinear observation scenarios, IEnSF converges to the exact posterior within about three to five iterations \cite{IEnSF}. On a linear harmonic oscillator, where the true posterior is available in closed form via the Kalman filter, plain EnSF still shows a large KL divergence against that ground truth, proving that the bias residing in the posterior score's formula rather than in numerical error from the reverse-time SDE solver \cite{IEnSF}. On 1000D Lorenz-96 with nonlinear arctan observations at 50\% coverage, IEnSF also substantially outperforms both LETKF and plain EnSF, though under direct observations at 25\% coverage it only matches LETKF's performance rather than exceeding it \cite{IEnSF}. This confirms that EnSF's error is structural rather than numerical, that the fix generalizes from a diagnostic case to a high-dimensional chaotic system, and that its advantage over LETKF is specifically tied to the nonlinearity of the observation operator rather than holding unconditionally.

\subsection{Ensemble Flow Filter (EnFF)}
IEnSF eliminates the structural bias in EnSF's posterior score but at the cost of one full RSDE solve per iteration, and the Euler-Maruyama discretization of the reverse-time SDE carries $\mathcal{O}(\sqrt{\Delta\tau})$ error per step, forcing fine discretization regardless of score accuracy and making each assimilation cycle slow for long-trajectory operational DA \cite{EnFF}.

The key insight of the Ensemble Flow Filter (EnFF) is that EnSF's RSDE and an ODE-based flow are theoretically interchangeable under flow matching, allowing safely larger integration steps for the same accuracy by lowering discretization error from $\mathcal{O}(\sqrt{\Delta\tau})$ to $\mathcal{O}(\Delta\tau)$ \cite{EnFF}. EnFF preserves EnSF's training-free Monte Carlo estimation principle, but estimates a marginal vector field instead of a score function, and follows a deterministic flow instead of a stochastic reverse-time process. Two variants realize this framework, EnFF-OT matching EnSF's standard Gaussian reference for speed alone, and EnFF-F2P exploiting flow matching's flexibility to use a problem-specific reference instead \cite{EnFF}.

The marginal vector field transporting the reference distribution to the prior is estimated through the same training-free Monte Carlo approach as EnSF's score estimator \cite{EnFF}:
\begin{equation} \label{eq:enff_vf}
    u_\tau(z) \approx \sum_{n=1}^{N} w_n(z)\, 
    u_\tau(z \mid z_0^{(n)}, z_1^{(n)}), \quad 
    w_n(z) = \frac{p_\tau(z \mid z_0^{(n)}, 
    z_1^{(n)})}{\sum_{m=1}^{N} p_\tau(z \mid 
    z_0^{(m)}, z_1^{(m)})},
\end{equation}
where $w_n(z)$ is the normalized weight measuring how compatible reference sample $z_0^{(n)}$ is with the current point $z$, $z_1^{(n)}$ are prior ensemble members, and $z_0^{(n)}$ are reference samples. EnFF-F2P is the primary contribution: it discards the standard Gaussian reference entirely and sets the reference to the previous posterior distribution instead, coupling reference and prior through the physical transition kernel $\rho(z_0, z_1) := p(z_1 \mid z_0)\rho_0(z_0)$ \cite{EnFF}. The conditional vector field becomes $u_\tau(z \mid z_0, z_1) = z_1 - z_0$ with straight-line flow $\phi_\tau = z + \tau(z_1 - z_0)$. Because the previous filtering distribution is already close to the current prior through the physical dynamics, F2P trajectories are substantially straighter than OT trajectories, requiring fewer ODE steps for equivalent accuracy \cite{EnFF}. Observations are incorporated through a localized guidance term that linearizes the likelihood at the conditional mean of $z_1$ given the current point, avoiding the degeneracy that importance-weighted guidance can produce when the likelihood is sharply peaked \cite{EnFF}.

Across Lorenz-96, the Kuramoto-Sivashinsky PDE ($d=1024$), and 2D Navier-Stokes ($256\times256$), both EnFF variants are consistently more accurate than EnSF at every tested sampling step count from $T=5$ to $T=100$, with widen gaps under low $T$ \cite{EnFF}. For Navier-Stokes specifically, EnSF fails to run stably at all below $T=10$. EnFF-F2P's optimal hyperparameters also vary far less with $T$ than EnSF's, meaning a cheap search at $T=5$ transfers reliably to larger step counts \cite{EnFF}. Against classical baselines, LETKF remains the most accurate method overall but is roughly two orders of magnitude slower than the EnFF variants \cite{EnFF}. EnFF only overtakes LETKF on accuracy at the largest tested grid, Navier-Stokes at $256\times256$; EnKF-PO, by contrast, fails to converge at that same largest setting \cite{EnFF}. The result is a genuine cost-accuracy tradeoff rather than a clean sweep, EnFF trades a small amount of LETKF's accuracy for a large and widening efficiency advantage as the problem gets harder.

\subsection{Two-Phase Flow EnSF}
IEnSF and EnFF repair EnSF's own mechanics, but both are still tested on the same canonical stress-test systems. Act I asks whether that mechanism needs to change at all once it faces a harder demand, and two-phase flow in porous media is its first test: a specific engineering domain, oil and gas recovery, geothermal energy, CO$_2$ sequestration, where the state couples multiple physical fields together and the forward model is built on an unknown, spatially heterogeneous permeability field \cite{hu2025ensemble}.

The key insight of Two-Phase Flow EnSF is that a single score function and a single RSDE update can carry the correction across multiple physical fields at once, by stacking saturation, velocity, and pressure into one joint state vector, $X_{t_n} := \left(s_h^n, \mathbf{u}_h^n, p_h^n\right)$, rather than requiring a separate filter per field \cite{hu2025ensemble}. The forward dynamics are supplied by a mixed finite element solver for the two-phase flow system, advanced with an Implicit-Pressure-Explicit-Saturation (IMPES) scheme \cite{hu2025ensemble}:
\begin{align}
    \mathbf{u}^{n+1} + \mathbf{K}\lambda(s^n)\nabla p^{n+1} &= 0 \label{eq:twophase_impes_a} \\
    \nabla \cdot \mathbf{u}^{n+1} &= q(t^{n+1}) \label{eq:twophase_impes_b} \\
    \frac{s^{n+1} - s^n}{\delta t} + \mathbf{u}^{n+1} \cdot \nabla F(s^n) &= 0 \label{eq:twophase_impes_c}
\end{align}
Given the previous saturation $s^n$, the pressure $p^{n+1}$ and velocity $\mathbf{u}^{n+1}$ are solved implicitly from \eqref{eq:twophase_impes_a}-\eqref{eq:twophase_impes_b}, and this velocity field is then substituted into \eqref{eq:twophase_impes_c} to update the saturation explicitly. This system depends on the permeability field $\mathbf{K}$, a quantity that is never directly known in practice and must instead be assumed, so the filter must correct for a forward model built on uncertain permeability, not just for noisy observations of a correct one \cite{hu2025ensemble}.

The method is tested in three scenarios on a $16{,}512$-dimensional state, each presenting a different kind of gap in what the forward model knows about the permeability field it depends on \cite{hu2025ensemble}. With widespread uncertainty in the permeability, EnSF alone drives the saturation RMSE from over $0.35$ with no observation down to roughly $0.02$ at full observation \cite{hu2025ensemble}. Against LETKF, whose linear update cannot properly exploit the nonlinear share of the added observations, the same pattern holds across both remaining scenarios, including a fracture network the forward model has no knowledge of: the two methods are indistinguishable at $10\%$ observation, but EnSF pulls decisively ahead as coverage increases, both in accuracy and in cost, reaching more than a $10\times$ speedup over LETKF at full observation \cite{hu2025ensemble}.

\subsection{Adaptive Learning EnSF for SPDEs}
Two-Phase Flow EnSF establishes that the mechanism survives contact with a specific engineering domain rather than a canonical stress-test system. Adaptive Learning EnSF for SPDEs pushes the same claim along the other axis of Act I, replacing one hardcoded forward solver with a general SPDE solver interface, and replacing a single fixed unknown like permeability with a disturbance that keeps perturbing the state at every step \cite{huynh2025scorebaseddiffusionmodelapproach}.

The key insight of Adaptive Learning EnSF for SPDEs is that the state can evolve under continuous stochastic forcing, a perturbation injected fresh at every step, rather than a single fixed unknown like permeability, and the recipe still holds across three physically unrelated PDE families \cite{huynh2025scorebaseddiffusionmodelapproach}. The general state model is an SPDE driven by a space-time Wiener process, realized in each numerical experiment as \cite{huynh2025scorebaseddiffusionmodelapproach}
\begin{equation} \label{eq:spde_forward}
    u_{n+1} = \mathbb{F}(u_n + \sigma_n \tilde{W}_{t_n}, t_n, \tilde{\omega}), \qquad \tilde{W}_{t_n} \sim N(0, \mathbf{I}_d),
\end{equation}
where $\tilde\omega$ represents unknown coefficients or parameters specific to the PDE at hand, and $\sigma_n \tilde{W}_{t_n}$ is a per-step Gaussian perturbation folded into the current forecast before the solver advances it, so each ensemble member follows its own stochastic trajectory rather than the deterministic path of the reference solution \cite{huynh2025scorebaseddiffusionmodelapproach}. Leaving $\mathbb{F}$ unspecified does not mean solver choice is incidental: for Burgers' equation, a TVB slope limiter is required alongside the discontinuous-Galerkin scheme to prevent noise accumulation; for Navier-Stokes, the Lagrange-multiplier scheme is chosen to preserve energy dissipation; for Allen-Cahn, the BDF2 scheme is chosen to preserve the maximum bound principle and unconditional energy stability \cite{huynh2025scorebaseddiffusionmodelapproach}.

The method is tested across three PDE families, Burgers' equation, Navier-Stokes in two configurations (the Taylor-Green vortex and an external-force-driven flow), and Allen-Cahn, each compared against LETKF at its sparsest tested observation level: $10\%$, $7\%$, $20\%$, and $10\%$ respectively \cite{huynh2025scorebaseddiffusionmodelapproach}. In every configuration the same shape holds on both axes that matter: EnSF's RMSE settles onto a low, stable plateau, roughly $0.07$ to $0.08$ for Burgers and Taylor-Green, $0.05$ to $0.13$ for the force-driven flow and Allen-Cahn, while LETKF stabilizes substantially higher, at roughly $0.12$, $0.26$, $0.25$, and $0.69$ \cite{huynh2025scorebaseddiffusionmodelapproach}. EnSF's wall-clock cost per step is also at or below LETKF's throughout, from a wide margin on Burgers and Allen-Cahn to a narrow one on Taylor-Green \cite{huynh2025scorebaseddiffusionmodelapproach}. Each RMSE is only meaningful within its own system, since the four PDEs measure different physical quantities on different scales. What generalizes is the pattern: EnSF converges low and cheap, LETKF converges higher and costlier. Combined with Two-Phase Flow, this confirms Act I's claim on a second axis that a general solver survives contact with any SPDE, not just porous media.

\subsection{Fire-EnSF}
Two-Phase Flow EnSF coupled multiple physical fields into one state, and Adaptive Learning EnSF for SPDEs generalized the forward operator to any solver an SPDE requires, but in both cases the state's own dimension stayed fixed throughout \cite{hu2025ensemble, huynh2025scorebaseddiffusionmodelapproach}. Fire-EnSF tests Act I's claim along a third axis, whether the state itself has to stay the same size at all. For wildfire, the fire perimeter is a polygon whose vertex count grows continuously as the burning area expands, so the state vector $\mathbf{x}_t \in \mathbb{R}^{2n_t}$ has a dimension $2n_t$ that increases at every assimilation step \cite{Fire-EnSF}.

Fire-EnSF's key insight is that EnSF's general strengths match wildfire's specific demands: no large sample sets as dimension grows into the thousands, no training delay for hourly forecasts, and native handling of both dense remote sensing and sparse UAV coverage \cite{Fire-EnSF}. The changing vertex count is handled separately, by re-interpolating every ensemble member to a shared dimension before the score update runs \cite{Fire-EnSF}.

FARSITE is an operational fire growth simulation model driven by topography, fuel characteristics, and weather conditions \cite{Fire-EnSF}. After it propagates the ensemble, each member and the observed perimeter carry different vertex counts, creating a dimensional mismatch that must be resolved before the score update \cite{Fire-EnSF}. The first preprocessing step identifies the maximum vertex count across all ensemble members and the observation \cite{Fire-EnSF}:
\begin{equation} \label{eq:reinterp}
    n_{t+1|t} = \max\!\left(
    n_{t+1|t}^1, \ldots, n_{t+1|t}^N, 
    m_{t+1}\right),
\end{equation}
where $n_{t+1|t}^i$ is the vertex count of the $i$-th ensemble member after FARSITE propagation and $m_{t+1}$ is the vertex count of the observed perimeter. The vertices of all state and observation vectors are then re-interpolated on the 2D spatial domain to this common dimension, producing a unified ensemble where every member and the observation share the same state space \cite{Fire-EnSF}.

Under controlled simulation with known ground truth, EnSF achieves substantially lower RMSE than EnKF across all filtering steps while delivering a $5.4\times$ reduction in wall-clock time, with mean execution times of $0.672 \pm 0.012$ seconds versus $3.634 \pm 0.013$ seconds on an Intel Xeon CPU and NVIDIA RTX A6000 GPU, establishing that the accuracy advantage holds when the true state is measurable \cite{Fire-EnSF}. On four real California wildfires from the CECS dataset using VIIRS satellite-derived perimeters at 0.5-day intervals, standalone FARSITE deviates significantly from observations across all four cases \cite{Fire-EnSF}. EnSF reduces perimeter error and maintains geometrically valid polygons throughout every assimilation step \cite{Fire-EnSF}. EnKF fails to reduce error and produces broken polygon geometry because its estimation errors break the Hamiltonian circuit ordering that FARSITE's sequential vertex propagation requires, a structural failure that also drives its computational cost to at least two days per fire compared to Fire-EnSF's one hour \cite{Fire-EnSF}.

\subsection{United Filter}
Fire-EnSF fixes parameters like fuel moisture at constant, known values \cite{Fire-EnSF}. United Filter addresses the harder case: a parameter like permeability that is unknown. A wrong guess corrupts the state estimate, and a corrupted state gives no way to fix the guess, so state and parameter must be estimated together, without modifying EnSF's own score update \cite{UF}. The standard approach, the Augmented Ensemble Kalman Filter (AugEnKF), folds parameters into the state vector and updates both with one EnKF step \cite{UF}. Its accuracy depends entirely on EnKF's own performance, and its parameter estimates suffer because state estimation dominates the update \cite{UF}.

The Direct Filter needs a complete state observation to estimate parameters, but this problem only offers sparse coverage \cite{UF}. The key insight of the United Filter is that EnSF uses its full state estimate reconstructed from partial data to fill that gap \cite{UF}. The two are coupled in a loop, EnSF's state estimate feeds the Direct Filter's parameter estimate, and that parameter estimate feeds EnSF's next forward propagation \cite{UF}.
 
At iteration $r$ within assimilation step $t{+}1$, Stage I propagates the current ensemble through the state dynamics using the current parameter estimate $\bar{\theta}_{t+1}^{(r)}$ and applies the standard EnSF score update to obtain a refined state estimate $\bar{X}_{t+1}^{(r)}$ \cite{UF}. Stage II then performs parameter estimation through the Direct Filter, substituting EnSF's estimate $\bar{X}_{t+1}^{(r)}$ for the true state its likelihood formula ordinarily requires:
\begin{equation} \label{eq:direct_filter_likelihood}
    p\big(\bar{X}_{t+1}^{(r)} \mid \tilde{\theta}_{t+1}^{m,(r+1)}\big) \propto 
    \exp\!\left[-\frac{1}{2}\Big(\bar{X}_{t+1}^{\tilde{\theta}_{t+1}^{m,(r+1)}} - \bar{X}_{t+1}^{(r)}\Big)^T \Theta^{-1} \Big(\bar{X}_{t+1}^{\tilde{\theta}_{t+1}^{m,(r+1)}} - \bar{X}_{t+1}^{(r)}\Big)\right],
\end{equation}
where $\Theta$ is the covariance of the model uncertainty and $\bar{X}_{t+1}^{\tilde{\theta}_{t+1}^{m,(r+1)}}$ is the state predicted using parameter particle $\tilde{\theta}_{t+1}^{m,(r+1)}$ \cite{UF}. The weighted particles are resampled and averaged to produce an updated parameter estimate $\bar{\theta}_{t+1}^{(r+1)}$, which becomes the input to Stage I in the next iteration. After $R$ iterations, the final state and parameter estimates are taken as $\bar{X}_{t+1} := \bar{X}_{t+1}^{(R)}$ and $\bar{\theta}_{t+1} := \bar{\theta}_{t+1}^{(R)}$ \cite{UF}.

On a reduced fracture model for subsurface flow, tested under full, 75\%, and 50\% sparse observations with half transformed through the arctangent operator, the United Filter accurately recovers both the pressure and velocity fields and converges to the true permeability parameters across all three scenarios \cite{UF}. Under the same conditions, AugEnKF recovers only a noisy fraction of the pressure field, produces chaotic velocity estimates, and its parameter particles fail to converge, instead oscillating between the two boundary values enforced to prevent divergence \cite{UF}. On a harder version of the same model with anisotropic permeability and three noise levels, and separately on an advection-diffusion transport case, the United Filter stays accurate and largely insensitive to noise, while AugEnKF is entirely incorrect under anisotropy and fails outright once noise increases in transport, closing Act I with the same pattern as Fire-EnSF: EnSF's own score update never had to change \cite{UF}.

\subsection{Score-DA}
Act I's forward models were always physics, consistent by construction no matter the input. Act II tests a genuinely different kind of forward model: a machine learning forecaster, competent only within what it was trained on. At deployment it must run recursively, feeding its own past prediction back in \cite{tang2026score}. Errors compound as the forecast drifts onto inputs it never learned, regardless of architecture or training \cite{tang2026score}.

The key insight of Score-DA is that this recursive error accumulation can be corrected without redesigning or retraining the ML model \cite{tang2026score}. At each assimilation step, the ML model is run once per ensemble member, producing an ensemble of forecasts that serves as EnSF's prior. The standard posterior score update is applied exactly as it would be with a physics-based forward model, using the true observational data to correct the ensemble before it becomes the input for the next recursive step \cite{tang2026score}.

Score-DA distinguishes two recursive prediction modes for a trained forecasting model $f(\cdot)$, demonstrated here on LSTM and R-DeepONet \cite{tang2026score}. In single-step prediction (SSP), each forecast is conditioned on the true past states, so errors cannot accumulate. In long-term prediction (LTP), each forecast is conditioned on the model's own previous predictions \cite{tang2026score}:
\begin{equation} \label{eq:ltp}
    (\text{LTP})_k = f\big(\{(\text{LTP})_i\}_{i=k-K}^{k-1}\big),
\end{equation}
where $K$ is the length of the input window and $(\text{LTP})_i = Y_i$ for the first $K$ steps before recursion begins. Because real deployment lacks true future states, LTP is what any operational use of $f(\cdot)$ must run, and its compounding error is the problem Score-DA targets. At each assimilation cycle, the LTP forecast is corrected using EnSF. The prior score is estimated from the ensemble of LTP predictions, the posterior score incorporates the new observation, and the RSDE produces a corrected ensemble that replaces the raw LTP output before the next recursive step \cite{tang2026score}.

On the 20-dimensional Lorenz-96 system with an LSTM as the LTP forecaster and arctan observations, EnSF-corrected LTP tracks the true trajectory closely under both full and sparse observation coverage, while EnKF shows visibly larger oscillations under full observations and is outperformed by EnSF under sparse observations \cite{tang2026score}. The advantage grows sharply when the underlying LSTM is poorly trained. On models trained with either insufficient or noisy data, EnKF becomes unstable and produces spuriously large deviations from the true state, while EnSF continues to track the trajectory reliably, though with more oscillation under the weaker model \cite{tang2026score}. On the Korteweg-De Vries equation with an R-DeepONet forecaster, the same pattern holds: under partial observations covering only half the spatial grid, EnKF's error is comparable to using no assimilation at all, while EnSF continues to substantially reduce the error accumulated by the LTP rollout \cite{tang2026score}. Across every tested condition, from a well-trained model with full observations to a poorly trained model with only partial observations, EnSF consistently corrects recursive drift that neither the raw LTP forecast nor EnKF can control.

\subsection{EnSF for Real-Data Energy Consumption Forecast Correction}
Score-DA shows that EnSF can correct a pretrained neural forecaster's drift, but only on synthetic dynamical systems built for research, where the ``sensor data" is clean output taken directly from a known equation \cite{tang2026score}. This subsection tests Act II's claim on the more challenging case: real deployed forecasters trained on messy, real-world measurements.

This paper's key insight is that the correction mechanism itself needs no adaptation to make that jump: the same unmodified EnSF update that corrects a synthetic forecaster's drift applies directly to STLLM, a pretrained attention-based spatio-temporal model forecasting real households' energy use, treating the network as a fixed black box exactly as Score-DA treats its LSTM \cite{hu2026diffusion, tang2026score}.

Because this forecasting model conditions on a window of past states rather than a single previous one, each ensemble member must carry its own window forward \cite{hu2026diffusion},
\begin{equation} \label{eq:window}
    W_n^{(m)} = \left(z_{n-T+1|n-T+1}^{(m)}, \dots, z_{n|n}^{(m)}\right), \qquad
    z_{n+1|n}^{(m)} = f_\theta\left(W_n^{(m)}, \eta_{t_n}^{(m)}\right),
\end{equation} 
where $f_\theta$ is the pretrained forecasting model, with $\theta$ never updated by the filter. After the first assimilation step, this window is populated entirely by previously assimilated states rather than the true trajectory, so the model is driven recursively by its own corrected output rather than by ground truth \cite{hu2026diffusion}. The only genuinely new mechanism here is this windowed bookkeeping, EnSF's own score estimator and posterior update never change \cite{hu2026diffusion}.

The method is evaluated on real hourly consumption data from $5{,}000$ utility users, using forecasting models that condition on one, four, or twelve prior time steps \cite{hu2026diffusion}. Under direct prediction, with the true input supplied at every step, the twelve-step model wins, reaching an RMSE of $0.52$ \cite{hu2026diffusion}. Forced to run open-loop instead, using its own output as input, that same model becomes the worst of the three, with RMSE rising to $2.24$ \cite{hu2026diffusion}. This reversal, the best model in isolation becoming the least reliable once it runs on its own predictions, is the paper's clearest case for why data assimilation is needed at all \cite{hu2026diffusion}. EnSF is then applied to the four-step model under this open-loop setting \cite{hu2026diffusion}. It substantially reduces RMSE whether that model is insufficiently trained or fully trained, improves further as observation coverage rises, and in a follow-up comparison against EnKF under mixed direct and nonlinear observations, sustains a lower RMSE throughout \cite{hu2026diffusion}. 

\subsection{Linear Regression EnSF (EnSF-LR)}
Score-DA and Energy-Consumption Forecast Correction both show Act II's restraint holding against a machine learning forecaster whose errors compound over time, corrected without touching the filter's own equations \cite{hu2026diffusion, tang2026score}. That restraint runs out when it comes to sparse observation. EnKF has always spread an observed correction to unobserved components through ensemble covariance, but EnSF never had that mechanism \cite{xiong2026twostepensemblescorefilter}.

The key insight of EnSF-LR is that the correction for unobserved components doesn't need deriving; it can be read directly off the ensemble as a sample covariance similarly to EnKF \cite{xiong2026twostepensemblescorefilter}. EnSF's own nonlinear update on the observed components stays unchanged. The sample covariance makes no assumption about the shape of the density, and costs almost nothing since the ensemble already exists. No training is required anywhere in this fix \cite{xiong2026twostepensemblescorefilter}.

For each ensemble member, EnSF computes an analysis increment on the observed components as usual, the difference between its EnSF posterior and prior. EnSF-LR then maps this increment onto the unobserved components through a linear regression step \cite{xiong2026twostepensemblescorefilter},
\begin{equation} \label{eq:ensflr_regression}
    \Delta\mathbf{x}_t^{unobs,(i)} = \mathbf{P}_{uo}\mathbf{P}_{oo}^{-1}\Delta\mathbf{x}_t^{obs,(i)},
\end{equation}
where $\mathbf{P}_{oo}$ and $\mathbf{P}_{uo}$ are ensemble-estimated covariances between observed-observed and unobserved-observed components, projecting each member's observed-side correction onto the unobserved components in proportion to how strongly the two have historically covaried \cite{xiong2026twostepensemblescorefilter}.

On Lorenz-63 and Lorenz-96 with partially observed states, EnSF-LR is compared against both plain EnSF and EnKF under linear and nonlinear observation operators \cite{xiong2026twostepensemblescorefilter}. Under linear observations, EnSF-LR substantially narrows the gap to EnKF, though the residual gap widens for the higher-dimensional Lorenz-96 case, where the demands on ensemble-estimated cross-covariances are greater \cite{xiong2026twostepensemblescorefilter}. Under nonlinear observations, EnSF-LR reduces RMSE by up to 57-60\% relative to EnSF and 23-35\% relative to EnKF \cite{xiong2026twostepensemblescorefilter}. In the Lorenz-96 case specifically, this improvement is concentrated in better-reconstructed unobserved components, confirming the regression step itself is doing the work rather than riding on EnSF's existing strength on the observed side \cite{xiong2026twostepensemblescorefilter}. This is a structural consequence of computing the regression coefficients from EnSF's nonlinear increment rather than from a jointly optimal linear-Gaussian update. It confirms the hybrid earns its clearest advantage where EnSF's nonlinear correction has room to matter, while the borrowed regression step recovers unobserved components without any new observation-space mechanism.

\subsection{Inpainting-EnSF}
Any covariance-based fix like EnSF-LR depends on that covariance existing and being close enough to linear to regress against \cite{xiong2026twostepensemblescorefilter}. When coverage is so extreme that no covariance structure connects observed and unobserved components, or the true relationship between them is too nonlinear for a linear regression to capture, the correction has nothing to propagate through, and the posterior along those dimensions remains entirely untouched by the data.

Inpainting-EnSF's key insight is that a state with unobserved components is structurally the same problem as an image with missing pixels \cite{Inpainting}. The missing information is not random; it is spatially correlated with what remains, and can be reconstructed from that surrounding structure rather than from any observation-driven gradient \cite{Inpainting}. EnSF's score-based update is kept for the observed components, where the gradient signal is intact, and is replaced entirely by image inpainting for the unobserved components, where no such signal exists \cite{Inpainting}.

Three inpainting methods are incorporated, each exploiting spatial structure differently \cite{Inpainting}. Navier-Stokes inpainting treats the state field as a stream function for a two-dimensional incompressible flow, with the Laplacian playing the role of vorticity, and propagates isophote lines continuously from observed into unobserved regions by solving a vorticity transport equation \cite{Inpainting}:
\begin{equation} \label{eq:ns_inpainting}
    \frac{\partial \omega_{\text{NS}}}{\partial t} + 
    v \cdot \nabla \omega_{\text{NS}} = \nu \nabla \cdot 
    (\nabla \omega_{\text{NS}}), \quad v = \nabla^\perp 
    X^{\text{unobs}}, \quad 
    \Delta X^{\text{unobs}} = \omega_{\text{NS}}
\end{equation}
where $\nabla^\perp$ denotes the perpendicular gradient operator, and $\nu$ is the fluid viscosity in Navier-Stokes. This method propagates geometric structure and edge continuity smoothly. Biharmonic inpainting enforces a fourth-order smooth extension of the observed state into unobserved regions by solving \cite{Inpainting}
\begin{equation} \label{eq:biharmonic}
    \Delta^2 X^{\text{unobs}} = 0 
    \quad \text{on the missing region,}
\end{equation}
with boundary conditions matching the value and Laplacian of the observed state at the boundary \cite{Inpainting}. The Discrete Cosine Transform (DCT) dictionary method represents the state field as a sparse combination of predefined basis functions \cite{Inpainting}. The dictionary is adapted using posterior estimates from previous assimilation steps, retaining only the elements that best express the current state, making it particularly effective when the state exhibits recurring patterns or multiscale features that local PDE propagation would miss \cite{Inpainting}. Like EnSF-LR, none of these three methods require any network training. Their added cost is a per-step PDE solve or transform, not an upfront investment \cite{Inpainting}.

On the Surface Quasi-Geostrophic (SQG) model, Inpainting-EnSF substantially outperforms base EnSF and LETKF under arctan observations, $256 \times 256$ grids, 12-hour DA intervals, and $5\%$ observation coverage \cite{Inpainting}. LETKF's hyperparameters here are transferred from tuning at a coarser $64 \times 64$ resolution rather than tuned directly at this scale, since direct tuning at $256 \times 256$ was computationally prohibitive \cite{Inpainting}. At $5\%$ coverage, DCT outperforms both PDE methods because sparse boundary information is insufficient for local propagation, while DCT's global basis functions recover large-scale structure from the few available observations \cite{Inpainting}. At $25\%$ coverage, sufficient observed components allow PDE methods, particularly biharmonic, to reconstruct unobserved structure competitively with DCT \cite{Inpainting}.

\subsection{Latent-EnSF}
Spatial reconstruction, like Inpainting-EnSF, needs a geometry connecting observed and unobserved regions to reconstruct from \cite{Inpainting}. Without one, it has nothing to work with. Latent-EnSF avoids that requirement entirely by eliminating the observed and unobserved distinction before the score update runs, rather than recovering it afterward \cite{Latent-EnSF}.

The key insight of Latent-EnSF is that if both the full state and the sparse observation are compressed into the same low-dimensional latent space, the latent observation map simplifies to approximately the identity $\zeta_t \approx \xi_t + \hat{\gamma}_t$, where $\xi_t$ is the latent state, $\zeta_t$ is the latent observation, and $\hat{\gamma}_t$ is latent noise \cite{Latent-EnSF}. Every latent dimension is then effectively observed, so the likelihood gradient is nonzero throughout the latent space. Latent-EnSF preserves EnSF's training-free Monte Carlo score estimator, posterior score update, and RSDE entirely unmodified, replacing only the space in which they operate: $\xi_t \in \mathbb{R}^{2r}$ rather than $x_t \in \mathbb{R}^d$ \cite{Latent-EnSF}.

This is achieved through a coupled Variational Autoencoder (VAE) with two encoders. The state encoder $\mathcal{E}: \mathbb{R}^d \to \mathbb{R}^{2r}$ maps the full state $x_t$ to latent $\xi_t = (\mu_t, \sigma_t^2)$, and the observation encoder $\mathcal{E}_{\text{obs}}: \mathbb{R}^m \to \mathbb{R}^{2r}$ maps the sparse observation $y_t$ to latent $\zeta_t = (\mu_t^{\text{obs}}, (\sigma_t^{\text{obs}})^2)$, both into the same latent space with $2r \ll d$ \cite{Latent-EnSF}. The critical innovation is a latent matching term in the training loss that forces both encoders to align their latent distributions \cite{Latent-EnSF}:
\begin{equation} \label{eq:latent_matching}
    \ell_t(\theta) = 
    \underbrace{
    \|x_t - \mathcal{D}(z_t)\|_2^2 + 
    \|x_t - \mathcal{D}(z_t^{\text{obs}})\|_2^2
    }_{\text{reconstruction}}
    + \underbrace{
    \|\mu_t - \mu_t^{\text{obs}}\|_2^2 + 
    \|\Sigma_t - \Sigma_t^{\text{obs}}\|_2^2
    }_{\text{latent matching}}
    + \text{Kullback-Leibler Divergence}
\end{equation}
where $z_t \sim \xi_t$ and $z_t^{\text{obs}} \sim \zeta_t$ are samples drawn from the state and observation latent distributions. The latent matching term is what makes the identity approximation valid. Without it, both encoders produce accurate reconstructions but from segregated regions of the latent space, so $\zeta_t$ is no longer a reliable proxy for $\xi_t$ and the score update receives inconsistent information \cite{Latent-EnSF}. With it, both encoders are constrained to the same latent region, making the sparse observation a consistent surrogate for the full state. The decoder $\mathcal{D}$ then reconstructs the full state from the posterior latent samples after the RSDE completes \cite{Latent-EnSF}.

On the shallow water model with a 225x observation reduction (0.44\% coverage), Latent-EnSF achieves comparable accuracy to EnSF at full observation, while base EnSF shows almost no improvement over a non-assimilated trajectory at this sparsity level \cite{Latent-EnSF}. On ERA5 geopotential height at 500 hPa (Z500) with $64\times$ spatial sparsity and once-daily assimilation, Latent-EnSF quickly levels off its error and holds that accuracy throughout the full 41-day evaluation window, while EnSF's assimilation error tracks the same curve as running no assimilation at all, providing essentially no correction \cite{Latent-EnSF}. These results confirm that the latent identity approximation eliminates the vanishing gradient problem even under extreme sparsity where no spatial reconstruction approach can operate, though the full-space physics simulator is still run to propagate the state before each encoding step, a cost this fix does not remove.

\subsection{Latent Dynamics EnSF (LD-EnSF)}
That remaining cost is the first thing LD-EnSF removes \cite{Latent-EnSF, LD-EnSF}. A second limitation emerges from the latent representation itself: Latent-EnSF's VAE produces oscillatory and non-smooth latent trajectories, making it difficult to learn stable surrogate dynamics in the latent space \cite{LD-EnSF}.

The key insight of LD-EnSF is that smooth latent trajectories make surrogate dynamics learnable, eliminating the need for a full-space simulator \cite{LD-EnSF}. On the state side, a single LDNet replaces both Latent-EnSF's VAE encoder and its external forward simulator, evolving the latent state directly \cite{LD-EnSF}. On the observation side, an LSTM replaces the VAE encoder instead. EnSF's score-based representation and the shared latent space are otherwise unchanged \cite{LD-EnSF}.

Three components operate entirely in the latent space. The LDNet consists of a dynamics network $\mathcal{F}_{\theta_1}$ that evolves the latent state $s_t \in \mathbb{R}^{d_s}$ conditioned on an uncertain parameter $u_t \in \mathbb{R}^{d_u}$, which captures unresolved physical variability such as Reynolds number or forcing terms \cite{LD-EnSF}:
\begin{equation} \label{eq:ldnet}
    s_t = s_{t-1} + \Delta t \, 
    \mathcal{F}_{\theta_1}(s_{t-1}, u_t),
\end{equation}
and a reconstruction network $\mathcal{R}_{\theta_2}$ that maps the latent state back to the full state at any spatial query point, called only once at the end of the assimilation cycle rather than at every step. A long short-term memory (LSTM) observation encoder maps the sparse observation history $y_{1:t}$ to a latent pair $(\hat{s}_t, \hat{u}_t)$, exploiting temporal correlations across observation history and handling irregular observation locations that Latent-EnSF's VAE encoder cannot accommodate \cite{LD-EnSF}. EnSF then runs in the augmented latent space $\kappa_t = (s_t, u_t)$ with an identity latent observation map, so every latent dimension is observed and the likelihood gradient is nonzero everywhere \cite{LD-EnSF}.

The original paper first evaluates LDNet as a standalone surrogate, tested on all three systems used later for assimilation \cite{LD-EnSF}. VAE-dyn, a VAE paired with a separate LSTM for propagation, since the VAE has no built-in dynamics, reaches relative RMSE of $0.964$, $1.33$, and $0.483$ on Kolmogorov, tsunami, and atmospheric respectively, indicating near-complete failure on the first two \cite{LD-EnSF}. LDNet, whose own dynamics network needs no external LSTM, cuts this to $0.0123$, $0.0168$, and $0.0656$ \cite{LD-EnSF}. The gap traces to VAE's oscillatory latent states causing the LSTM to accumulate error, versus LDNet's smooth trajectories \cite{LD-EnSF}.

With the surrogate validated, LD-EnSF is evaluated on three DA benchmarks of increasing complexity: Kolmogorov flow ($150 \times 150$, $0.44\%$ observations), tsunami modeling ($150 \times 150$, $0.44\%$), and atmospheric modeling ($512 \times 256$, $0.1\%$ spatial and $0.2\%$ temporal), each also tested with irregularly sampled locations to confirm the advantage isn't limited to fixed grids \cite{LD-EnSF}. LD-EnSF achieves the lowest error across all three while delivering speedups of ${\approx}2\times10^5\times$, ${\approx}4\times10^3\times$, and ${\approx}5\times10^5\times$ over full-space methods, paid for with 12 to 22 hours of offline training \cite{LD-EnSF}. EnSF fails at this sparsity, LETKF diverges outright in the atmospheric case and is excluded entirely, and Latent-EnSF, while handling sparsity, loses accuracy to its own oscillatory latent states \cite{LD-EnSF}.

\subsection{Diffusion-Based Uncertainty Quantification (DBUQ)}
Every method so far has kept its own filtering step training-free and assumed the forward model can be run many times within one assimilation cycle. DBUQ breaks both \cite{lu2024diffusion}. It also leaves sequential filtering behind entirely: it estimates a fixed set of unknown parameters from the full data set at once, not a state that changes over time and gets corrected step by step \cite{lu2024diffusion}. Earth system models make repeated forward-model runs especially costly: a single land-surface simulation can take hours, and Markov Chain Monte Carlo, the standard calibration tool, often needs hundreds of thousands of such runs to converge \cite{lu2024diffusion}. At that cost, calibrating a model site by site across every climate zone is effectively out of reach.

The key insight of DBUQ is that the expensive part of a diffusion model is not sampling from the posterior, but solving the reverse-time process at all \cite{lu2024diffusion}. If that process can instead be learned once as a reusable function, every later observation is answered by a single evaluation of that function rather than a fresh solve \cite{lu2024diffusion}.

DBUQ estimates the score function through the same importance-weighted Monte Carlo approach as EnSF, using a mini-batch of prior samples $\{x^j, y^j\}_{j=1}^{J}$ generated by running the physical model $\mathcal{M}(\cdot)$ at each prior parameter sample \cite{lu2024diffusion}. With this estimated score, DBUQ solves the reverse-time ordinary differential equation \cite{lu2024diffusion}
\begin{equation} \label{eq:dbuq_ode}
    \frac{dZ_\tau}{d\tau} = b(\tau)Z_\tau - \frac{1}{2}\sigma^2(\tau)S(Z_\tau, \tau)
\end{equation}
from $\tau=1$ to $\tau=0$, in place of EnSF's RSDE. Because this equation has no stochastic term, the same Gaussian sample $Z_1$ always maps to the same target sample $Z_0 = X|Y$, giving a smooth, repeatable relationship that the SDE's random path cannot provide. DBUQ solves this ODE once per prior sample to build a labeled data set $\{(x^m, y^m, z^m)\}_{m=1}^{M}$, then trains a feedforward network $F(Y,Z;\theta)$ on these pairs via standard supervised learning \cite{lu2024diffusion}. Once trained, a new observation $y$ requires only evaluating $F(y,Z;\theta)$ at standard Gaussian samples $Z$ to produce the corresponding posterior samples, with no further score estimation or ODE solving needed \cite{lu2024diffusion}.

On a one-dimensional bimodal test problem, DBUQ recovers both modes of the exact posterior with KL divergence of only $0.036$ and $0.055$ at two observations, confirming that the learned generator preserves the shape of the posterior and not just its mean \cite{lu2024diffusion}. DBUQ is then applied to calibrating eight parameters of the E3SM Land Model at the Missouri Ozark AmeriFlux site and compared with MCMC. Both methods start from the same 1,000 real ELM simulations \cite{lu2024diffusion}. MCMC uses them to build a fast surrogate to sample from, DBUQ uses them to build its training data. From that shared starting point, MCMC's sampling takes about 5 hours, while DBUQ's entire process, training plus generating samples, takes under 10 minutes \cite{lu2024diffusion}. This is the gap identified at the outset: DBUQ turns a per-site, hours-to-days calibration cost into a one-time training cost, making the global, site-specific calibration that direct MCMC forecloses computationally realistic.

\subsection{STORM}
DBUQ eliminates repeated iterative sampling for parameter calibration, but sequential state estimation across time still confronts a deeper structural barrier at planetary scale \cite{lu2024diffusion}. Conventional data assimilation, including every extension surveyed so far, alternates between a forecast step and an update step \cite{kalman1961new}. At kilometer-scale resolution, this two-step structure requires repeated global synchronization between the forecast and update modules every cycle, clashing with exascale GPU systems built for dense, independent computation rather than frequent whole-domain exchange \cite{wang2026global}.

STORM's key insight is to unify forecast and assimilation into one generative sampling process, conditioning directly on history and observations \cite{wang2026global}. The prior score is estimated by treating it as equivalent to denoising the Earth system state, though full attention across every token is infeasible at planetary scale, a limit the tiling scheme below resolves \cite{wang2026global}.

STORM keeps EnSF's posterior decomposition, $S_{\text{posterior}} \approx S_{\text{prior}} + S_{\text{likelihood}}$, but replaces EnSF's training-free Monte Carlo estimator with a trained network, applied tile by tile \cite{wang2026global}. Its main idea is that estimating the prior score is equivalent to denoising the Earth system state, made explicit via Tweedie's identity \cite{wang2026global},
\begin{equation} \label{eq:storm_prior}
    S_{\text{prior}}(z_t, \sigma_t \mid \mathbf{x}_{t-K:t-1}) \approx \frac{D_\theta(z_t, \sigma_t; \mathbf{x}_{t-K:t-1}) - z_t}{\sigma_t^2},
\end{equation}
where $D_\theta$ is a learned denoiser recovering the clean state from its noise-corrupted version, conditioned on $K$ historical frames. $D_\theta$ decouples space from time: historical frames are aggregated across both variable and temporal dimensions, the current noisy state only across variables, removing the multiplicative coupling behind TimeSformer's $\mathcal{O}(K^2N + KN^2)$ complexity and cutting it to $\mathcal{O}(N^2)$ per layer \cite{wang2026global}. To remove the remaining quadratic cost in spatial tokens $N$, the domain is split into overlapping tiles, with denoised outputs averaged in the overlaps at every step, propagating context tile to tile until global correlations emerge at linear cost \cite{wang2026global}.

On a 10M-parameter model, STORM reaches 196.6 million spatiotemporal tokens on 16 GPUs, against TimeSformer's ceiling of 3.1 million on the same hardware, roughly two orders of magnitude further before hitting the quadratic complexity limit \cite{wang2026global}. Ensemble generation time on this same model stays near-constant at $\sim$34 seconds as ensemble size grows from 64 to over 32,000 members, evidence that removing the forecast-correction synchronization bottleneck lets ensemble size scale independently of runtime \cite{wang2026global}. At full scale, 32,768 GPUs, STORM reaches 20 billion tokens with 177,000 temporal frames, a regime the paper states existing transformer-based methods cannot reach at all, sustaining 1.6 ExaFLOP/s at 61-64\% strong scaling efficiency \cite{wang2026global}. On four rapidly-intensifying hurricanes, one-stage DA cuts track uncertainty by up to 60\% at 20\% observation coverage, and on regional 5-km temperature prediction, 50\% coverage reduces RMSE by 2-3 K, correcting diurnal heating errors exceeding 10 K near the US-Mexico border \cite{wang2026global}. These accuracy gains confirm STORM's tiling is fast without sacrificing assimilation quality, though they test overall accuracy rather than whether tile-to-tile propagation specifically preserves long-range spatial correlation \cite{wang2026global}. Together with DBUQ, this closes Act IV: twelve of the fourteen extensions left EnSF's training-free update untouched, only these two gave it up, each for a scale neither could have reached otherwise.

\bibliographystyle{plain}
\bibliography{source}

@article{BDSDE,
  title={Data informed solution estimation for forward-backward stochastic differential equations},
  author={Bao, Feng and Cao, Yanzhao and Yong, Jiongmin},
  journal={Analysis and Applications},
  volume={19},
  number={03},
  pages={439--464},
  year={2021},
  publisher={World Scientific}
}

@article{FBSDE,
  title={A first order scheme for backward doubly stochastic differential equations},
  author={Bao, Feng and Cao, Yanzhao and Meir, Amnon and Zhao, Weidong},
  journal={SIAM/ASA Journal on Uncertainty Quantification},
  volume={4},
  number={1},
  pages={413--445},
  year={2016},
  publisher={SIAM}
}

@article{IEnSF,
  title={IEnSF: Iterative Ensemble Score Filter for Reducing Error in Posterior Score Estimation in Nonlinear Data Assimilation},
  author={Zhang, Zezhong and Bao, Feng and Zhang, Guannan},
  journal={arXiv preprint arXiv:2510.20159},
  year={2025}
}

@article{Inpainting,
  title={Ensemble score filter with image inpainting for data assimilation in tracking surface quasi-geostrophic dynamics with partial observations},
  author={Liang, Siming and Tran, Hoang and Bao, Feng and Chipilski, Hristo G and van Leeuwen, Peter Jan and Zhang, Guannan},
  journal={arXiv preprint arXiv:2501.12419},
  year={2025}
}

@inproceedings{Latent-EnSF,
  title={Latent-EnSF: A latent ensemble score filter for high-dimensional data assimilation with sparse observation data},
  author={Si, Phillip and Chen, Peng},
  booktitle={International Conference on Learning Representations},
  volume={2025},
  pages={94879--94895},
  year={2025}
}

@article{LD-EnSF,
  title={Ld-ensf: Synergizing latent dynamics with ensemble score filters for fast data assimilation with sparse observations},
  author={Xiao, Pengpeng and Si, Phillip and Chen, Peng},
  journal={arXiv preprint arXiv:2411.19305},
  year={2024}
}

@article{EnFF,
  title={Flow Matching-Based Generative Modeling for Efficient and Scalable Data Assimilation},
  author={Transue, Taos and Chen, Bohan and Takao, So and Wang, Bao},
  journal={arXiv e-prints},
  pages={arXiv--2508},
  year={2025}
}

@article{Fire-EnSF,
  title={Fire-EnSF: Wildfire Spread Data Assimilation using Ensemble Score Filter},
  author={Shi, Hongzheng and Wang, Yuhang and Liu, Xiao},
  journal={arXiv preprint arXiv:2510.15954},
  year={2025}
}

@article{bao2024ensemble,
  title={An ensemble score filter for tracking high-dimensional nonlinear dynamical systems},
  author={Bao, Feng and Zhang, Zezhong and Zhang, Guannan},
  journal={Computer Methods in Applied Mechanics and Engineering},
  volume={432},
  pages={117447},
  year={2024},
  publisher={Elsevier}
}

@article{UF,
  title={Joint state-parameter estimation for the reduced fracture model via the united filter},
  author={Huynh, Phuoc Toan and Zhang, Guannan and Bao, Feng and others},
  journal={Journal of Computational Physics},
  volume={538},
  pages={114159},
  year={2025},
  publisher={Elsevier}
}

@article{lu2024diffusion,
  title={A diffusion-based uncertainty quantification method to advance E3SM land model calibration},
  author={Lu, Dan and Liu, Yanfang and Zhang, Zezhong and Bao, Feng and Zhang, Guannan},
  journal={Journal of Geophysical Research: Machine Learning and Computation},
  volume={1},
  number={3},
  pages={e2024JH000234},
  year={2024},
  publisher={Wiley Online Library}
}

@article{wang2026global,
  title={Global Attention with Linear Complexity for Exascale Generative Data Assimilation in Earth System Prediction},
  author={Wang, Xiao and Zhang, Zezhong and Lyngaas, Isaac and Yoon, Hong-Jun and Choi, Jong-Youl and Liang, Siming and Wang, Janet and Chipilski, Hristo G and Aji, Ashwin M and Bao, Feng and others},
  journal={arXiv preprint arXiv:2604.16590},
  year={2026}
}

@article{kalman1960new,
  title={A new approach to linear filtering and prediction problems},
  author={Kalman, Rudolph Emil},
  year={1960}
}

@article{kalman1961new,
  title={New results in linear filtering and prediction theory},
  author={Kalman, Rudolph E and Bucy, Richard S},
  year={1961}
}

@inproceedings{gordon1993novel,
  title={Novel approach to nonlinear/non-Gaussian Bayesian state estimation},
  author={Gordon, Neil J and Salmond, David J and Smith, Adrian FM},
  booktitle={IEE proceedings F (radar and signal processing)},
  volume={140},
  number={2},
  pages={107--113},
  year={1993},
  organization={IET}
}

@article{evensen1994sequential,
  title={Sequential data assimilation with a nonlinear quasi-geostrophic model using Monte Carlo methods to forecast error statistics},
  author={Evensen, Geir},
  journal={Journal of Geophysical Research: Oceans},
  volume={99},
  number={C5},
  pages={10143--10162},
  year={1994},
  publisher={Wiley Online Library}
}

@article{snyder2008obstacles,
  title={Obstacles to high-dimensional particle filtering},
  author={Snyder, Chris and Bengtsson, Thomas and Bickel, Peter and Anderson, Jeff},
  journal={Monthly Weather Review},
  volume={136},
  number={12},
  pages={4629--4640},
  year={2008}
}

@article{zakai1969optimal,
  title={On the optimal filtering of diffusion processes},
  author={Zakai, Moshe},
  journal={Zeitschrift f{\"u}r Wahrscheinlichkeitstheorie und verwandte Gebiete},
  volume={11},
  number={3},
  pages={230--243},
  year={1969},
  publisher={Springer}
}

@article{pardoux1990adapted,
  title={Adapted solution of a backward stochastic differential equation},
  author={Pardoux, Etienne and Peng, Shige},
  journal={Systems \& control letters},
  volume={14},
  number={1},
  pages={55--61},
  year={1990},
  publisher={Elsevier}
}

@article{pardoux1994backward,
  title={Backward doubly stochastic differential equations and systems of quasilinear SPDEs},
  author={Pardoux, {\'E}tienne and Peng, Shige},
  journal={Probability theory and related fields},
  volume={98},
  number={2},
  pages={209--227},
  year={1994},
  publisher={Springer}
}

@article{bao2011numerical,
  title={Numerical solutions for forward backward doubly stochastic differential equations and zakai equations},
  author={Bao, Feng and Cao, Yanzhao and Zhao, Weidong},
  journal={Visualization of Mechanical Processes: An International Online Journal},
  volume={1},
  number={4},
  year={2011},
  publisher={Begel House Inc.}
}

@article{song2020score,
  title={Score-based generative modeling through stochastic differential equations},
  author={Song, Yang and Sohl-Dickstein, Jascha and Kingma, Diederik P and Kumar, Abhishek and Ermon, Stefano and Poole, Ben},
  journal={arXiv preprint arXiv:2011.13456},
  year={2020}
}

@incollection{stratonovich1965conditional,
  title={Conditional markov processes},
  author={Stratonovich, Ruslan Leont’evich},
  booktitle={Non-linear transformations of stochastic processes},
  pages={427--453},
  year={1965},
  publisher={Elsevier}
}

@article{kallianpur1968estimation,
  title={Estimation of stochastic systems: Arbitrary system process with additive white noise observation errors},
  author={Kallianpur, G and Striebel, Charlotte},
  journal={The Annals of Mathematical Statistics},
  volume={39},
  number={3},
  pages={785--801},
  year={1968},
  publisher={JSTOR}
}

@book{jazwinski2007stochastic,
  title={Stochastic processes and filtering theory},
  author={Jazwinski, Andrew H},
  year={2007},
  publisher={Courier Corporation}
}

@incollection{doucet2001introduction,
  title={An introduction to sequential Monte Carlo methods},
  author={Doucet, Arnaud and De Freitas, Nando and Gordon, Neil},
  booktitle={Sequential Monte Carlo methods in practice},
  pages={3--14},
  year={2001},
  publisher={Springer}
}

@article{bao2017adaptive,
  title={Adaptive meshfree backward SDE filter},
  author={Bao, Feng and Maroulas, Vasileios},
  journal={SIAM Journal on Scientific Computing},
  volume={39},
  number={6},
  pages={A2664--A2683},
  year={2017},
  publisher={SIAM}
}

@article{archibald2022kernel,
  title={Kernel learning backward SDE filter for data assimilation},
  author={Archibald, Richard and Bao, Feng},
  journal={Journal of Computational Physics},
  volume={455},
  pages={111009},
  year={2022},
  publisher={Elsevier}
}

@inproceedings{pardoux2005backward,
  title={Backward stochastic differential equations and quasilinear parabolic partial differential equations},
  author={Pardoux, Etienne and Peng, Shige},
  booktitle={Stochastic Partial Differential Equations and Their Applications: Proceedings of IFIP WG 7/1 International Conference University of North Carolina at Charlotte, NC June 6--8, 1991},
  pages={200--217},
  year={2005},
  organization={Springer}
}

@article{bao2024score,
  title={A score-based filter for nonlinear data assimilation},
  author={Bao, Feng and Zhang, Zezhong and Zhang, Guannan},
  journal={Journal of Computational Physics},
  volume={514},
  pages={113207},
  year={2024},
  publisher={Elsevier}
}

@article{anderson1982reverse,
  title={Reverse-time diffusion equation models},
  author={Anderson, Brian DO},
  journal={Stochastic Processes and their Applications},
  volume={12},
  number={3},
  pages={313--326},
  year={1982},
  publisher={Elsevier}
}

@article{kloeden1977numerical,
  title={The numerical solution of stochastic differential equations},
  author={Kloeden, Peter E and Pearson, RA},
  journal={The ANZIAM Journal},
  volume={20},
  number={1},
  pages={8--12},
  year={1977},
  publisher={Cambridge University Press}
}

@article{wang2026error,
  title={Error estimates of a training-free diffusion model for high-dimensional sampling},
  author={Wang, Pengjun and Zhang, Zezhong and Yang, Minglei and Bao, Feng and Cao, Yanzhao and Zhang, Guannan},
  journal={arXiv preprint arXiv:2601.19740},
  year={2026}
}

@inproceedings{julier1997new,
  title={New extension of the Kalman filter to nonlinear systems},
  author={Julier, Simon J and Uhlmann, Jeffrey K},
  booktitle={Signal processing, sensor fusion, and target recognition VI},
  volume={3068},
  pages={182--193},
  year={1997},
  organization={Spie}
}

@article{hunt2007efficient,
  title={Efficient data assimilation for spatiotemporal chaos: A local ensemble transform Kalman filter},
  author={Hunt, Brian R and Kostelich, Eric J and Szunyogh, Istvan},
  journal={Physica D: Nonlinear Phenomena},
  volume={230},
  number={1-2},
  pages={112--126},
  year={2007},
  publisher={Elsevier}
}

@article{tang2026score,
  title={A Score Filter Enhanced Data Assimilation Framework for Data-Driven Dynamical Systems},
  author={Tang, Jingqiao and Bausback, Ryan and Bao, Feng and Zhang, Guannan and Huynh, Phuoc-Toan},
  journal={arXiv preprint arXiv:2603.14863},
  year={2026}
}

@article{pitt1999filtering,
  title={Filtering via simulation: Auxiliary particle filters},
  author={Pitt, Michael K and Shephard, Neil},
  journal={Journal of the American statistical association},
  volume={94},
  number={446},
  pages={590--599},
  year={1999},
  publisher={Taylor \& Francis}
}

@incollection{murphy2001rao,
  title={Rao-Blackwellised particle filtering for dynamic Bayesian networks},
  author={Murphy, Kevin and Russell, Stuart},
  booktitle={Sequential Monte Carlo methods in practice},
  pages={499--515},
  year={2001},
  publisher={Springer}
}

@incollection{musso2001improving,
  title={Improving regularised particle filters},
  author={Musso, Christian and Oudjane, Nadia and Le Gland, Francois},
  booktitle={Sequential Monte Carlo methods in practice},
  pages={247--271},
  year={2001},
  publisher={Springer}
}

@misc{xiong2026twostepensemblescorefilter,
      title={A Two-Step Ensemble Score Filter for Data Assimilation in Partially Observed Systems}, 
      author={Zixiang Xiong and Feng Bao and Hristo G. Chipilski and Siming Liang and Jingqiao Tang and Guannan Zhang},
      year={2026},
      eprint={2606.28264},
      archivePrefix={arXiv},
      primaryClass={physics.ao-ph},
      url={https://arxiv.org/abs/2606.28264}, 
}

@article{hu2025ensemble,
  title={Ensemble score filter for data assimilation of two-phase flow models in porous media},
  author={Hu, Ruoyu and Poudel, Sanjeeb and Bao, Feng and Lee, Sanghyun},
  journal={Journal of Computational Physics},
  pages={114416},
  year={2025},
  publisher={Elsevier}
}

@misc{huynh2025scorebaseddiffusionmodelapproach,
      title={A Score-based Diffusion Model Approach for Adaptive Learning of Stochastic Partial Differential Equation Solutions}, 
      author={Toan Huynh and Ruth Lopez Fajardo and Guannan Zhang and Lili Ju and Feng Bao},
      year={2025},
      eprint={2508.06834},
      archivePrefix={arXiv},
      primaryClass={stat.CO},
      url={https://arxiv.org/abs/2508.06834}, 
}

@misc{hu2026diffusion,
      title={Diffusion Model-Based Data Assimilation for Real-World Energy Consumption Forecasting}, 
      author={Ruoyu Hu and Dahai Yu and Feng Bao and Guang Wang and Guannan Zhang},
      year={2026},
      eprint={2605.29072},
      archivePrefix={arXiv},
      primaryClass={cs.LG},
      url={https://arxiv.org/abs/2605.29072}, 
}

@article{bao2025nonlinear,
  title={Nonlinear ensemble filtering with diffusion models: Application to the surface quasigeostrophic dynamics},
  author={Bao, Feng and Chipilski, Hristo G and Liang, Siming and Zhang, Guannan and Whitaker, Jeffrey S},
  journal={Monthly Weather Review},
  volume={153},
  number={7},
  pages={1155--1169},
  year={2025},
  publisher={American Meteorological Society}
}

@article{xiong2026robustness,
  title={Robustness of the Ensemble Score Filter to the Type of Assimilated Observation Networks},
  author={Xiong, Zixiang and Liang, Siming and Bao, Feng and Zhang, Guannan and Chipilski, Hristo G},
  journal={Atmospheric Science Letters},
  volume={27},
  number={1},
  pages={e70004},
  year={2026},
  publisher={Wiley Online Library}
}

@article{bao2018backward,
  title={A backward doubly stochastic differential equation approach for nonlinear filtering problems},
  author={Bao, Feng and Cao, Yanzhao and Zhao, Weidong},
  journal={Communications in Computational Physics},
  volume={23},
  number={5},
  pages={1573--1601},
  year={2018}
}

@article{van2000unscented,
  title={The unscented particle filter},
  author={Van Der Merwe, Rudolph and Doucet, Arnaud and De Freitas, Nando and Wan, Eric},
  journal={Advances in neural information processing systems},
  volume={13},
  year={2000}
}

@article{whitaker2002ensemble,
  title={Ensemble data assimilation without perturbed observations},
  author={Whitaker, Jeffrey S and Hamill, Thomas M},
  journal={Monthly weather review},
  volume={130},
  number={7},
  pages={1913--1924},
  year={2002}
}

@article{hamill2001distance,
  title={Distance-dependent filtering of background error covariance estimates in an ensemble Kalman filter},
  author={Hamill, Thomas M and Whitaker, Jeffrey S and Snyder, Chris},
  journal={Monthly Weather Review},
  volume={129},
  number={11},
  pages={2776--2790},
  year={2001},
  publisher={American Meteorological Society}
}

@article{kurosawa2023statistical,
  title={A statistical hypothesis testing strategy for adaptively blending particle filters and ensemble Kalman filters for data assimilation},
  author={Kurosawa, Kenta and Poterjoy, Jonathan},
  journal={Monthly Weather Review},
  volume={151},
  number={1},
  pages={105--125},
  year={2023}
}

@article{andrieu2003efficient,
  title={Efficient particle filtering for jump Markov systems. Application to time-varying autoregressions},
  author={Andrieu, Christophe and Davy, Manuel and Doucet, Arnaud},
  journal={IEEE Transactions on signal processing},
  volume={51},
  number={7},
  pages={1762--1770},
  year={2003},
  publisher={IEEE}
}

@article{arulampalam2002tutorial,
  title={A tutorial on particle filters for online nonlinear/non-Gaussian Bayesian tracking},
  author={Arulampalam, M Sanjeev and Maskell, Simon and Gordon, Neil and Clapp, Tim},
  journal={IEEE Transactions on signal processing},
  volume={50},
  number={2},
  pages={174--188},
  year={2002},
  publisher={Ieee}
}

@inproceedings{lorenz1996predictability,
  title={Predictability: A problem partly solved},
  author={Lorenz, Edward N},
  booktitle={Proc. Seminar on predictability},
  volume={1},
  number={1},
  pages={1--18},
  year={1996},
  organization={Reading}
}

@article{carrassi2018data,
  title={Data assimilation in the geosciences: An overview of methods, issues, and perspectives},
  author={Carrassi, Alberto and Bocquet, Marc and Bertino, Laurent and Evensen, Geir},
  journal={Wiley Interdisciplinary Reviews: Climate Change},
  volume={9},
  number={5},
  pages={e535},
  year={2018},
  publisher={Wiley Online Library}
}

@article{xiong2025sensitivity,
  title={On the sensitivity of different ensemble filters to the type of assimilated observation networks},
  author={Xiong, Zixiang and Liang, Siming and Bao, Feng and Zhang, Guannan and Chipilski, Hristo G},
  journal={arXiv preprint arXiv:2505.04541},
  year={2025}
}

@article{liang2026covariance,
  title={Covariance-Informed Ensemble Score Filtering: Embedding Localized Covariance Geometry into Score-Based Transport},
  author={Liang, Siming and van Leeuwen, Peter Jan and Chipilski, Hristo G and Zhang, Zezhong and Bao, Feng and Zhang, Guannan},
  journal={arXiv preprint arXiv:2610.32148},
  year={2026}
}

@article{liang2026score,
  title={Score-Based Generative Data Assimilation for Integrating Aggregated Surveillance Data into Agent-Based Models in Epidemic Tracking},
  author={Liang, Siming and Hauck, Jacob and Yang, Minglei and Spannaus, Adam and Hanson, Heidi and Zhang, Guannan},
  journal={arXiv preprint arXiv:2609.01434},
  year={2026}
}

@article{zhang2025exact,
  title={Exact conditional score-guided generative modeling for amortized inference in uncertainty quantification},
  author={Zhang, Zezhong and Tatsuoka, Caroline and Xiu, Dongbin and Zhang, Guannan},
  journal={arXiv preprint arXiv:2506.18227},
  year={2025}
}
\end{document}